\documentclass[letterpaper]{article} 
\usepackage{aaai23}  
\usepackage{times}  
\usepackage{helvet}  
\usepackage{courier}  
\usepackage[hyphens]{url}  
\usepackage{graphicx} 
\usepackage{natbib}  
\usepackage{caption} 
\usepackage{algorithm}
\usepackage{algpseudocode}
\usepackage{booktabs}
\usepackage{macros}
\usepackage{paralist}
\usepackage{colonequals}
\usepackage{tabularx}
\usepackage{multirow}
\usepackage{tikz}
\usepackage{todonotes}
\usepackage{amsmath,amssymb} 
\usepackage{adjustbox}

\usepackage{pgfplots}
\usepackage{pgfplotstable} 

\pgfplotsset{
    node near coord/.style args={#1/#2/#3}{
        nodes near coords*={
            \ifnum\coordindex=#1 #2\fi
        },
        scatter/@pre marker code/.append code={
            \ifnum\coordindex=#1 \pgfplotsset{every node near coord/.append style=#3}\fi
        }
    },
    nodes near some coords/.style={ 
        scatter/@pre marker code/.code={},
        scatter/@post marker code/.code={},%
        node near coord/.list={#1} 
    }
}

\usepackage{newfloat}
\usepackage{listings}
\DeclareCaptionStyle{ruled}{labelfont=normalfont,labelsep=colon,strut=off} 
\floatstyle{ruled}
\newfloat{listing}{tb}{lst}{}
\floatname{listing}{Listing}
\title{Scalable Regularization of Scene Graph Generation Models using Symbolic Theories}
\author {
    Davide Buffelli\textsuperscript{\rm 1 $\dagger$},
    Efthymia Tsamoura \textsuperscript{\rm 2}
}
\affiliations {
    \textsuperscript{\rm 1} University of Padova\\
    \textsuperscript{\rm 2} Samsung AI Research
}

\begin{document}

\maketitle

Several techniques have recently aimed to improve the performance of deep learning models for Scene Graph Generation (SGG) by incorporating background knowledge.
State-of-the-art techniques can be divided into two families: one where the background knowledge is incorporated into the model in a subsymbolic fashion, and another in which the background knowledge is maintained in symbolic form. Despite promising results, both families of techniques face several shortcomings: the first one requires ad-hoc, more complex neural architectures increasing the training or inference cost; the second one suffers from limited scalability w.r.t. the size of the background knowledge. 
Our work introduces a regularization technique for injecting symbolic background knowledge into neural SGG models that overcomes the limitations of prior art. Our technique is model-agnostic, does not incur any cost at inference time, and scales to previously unmanageable background knowledge sizes.
We demonstrate that our technique can improve the accuracy of state-of-the-art SGG models, by up to 33\%.
\section{Introduction} \label{section:introduction}

\begin{figure}[t]
\centering
\includegraphics[width=\linewidth]{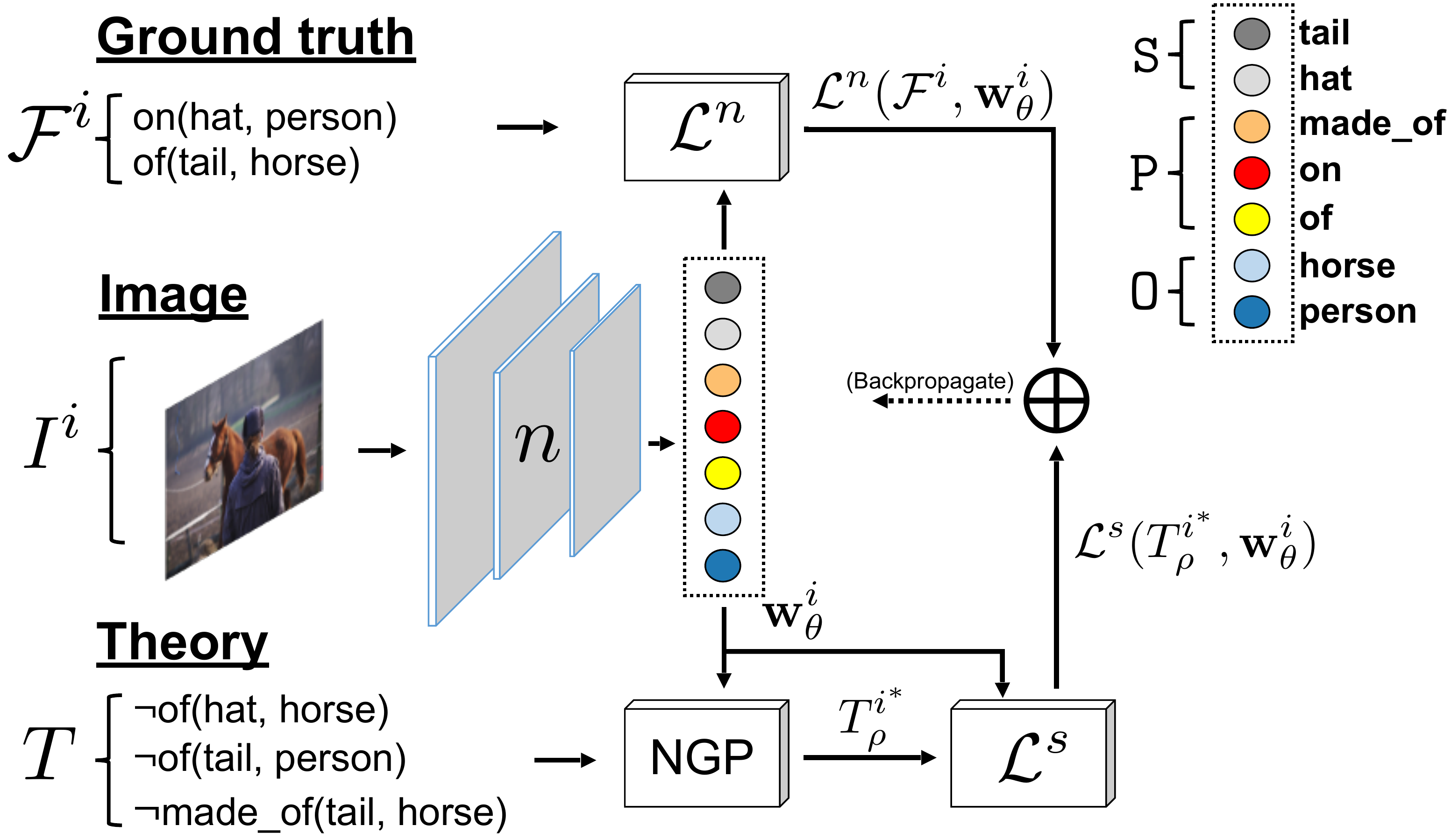} 
\caption{At training-time, background knowledge expressed through negative formulas in first-order logic is injected into a deep model $n$ so that the model’s predictions $\mathbf{w}_{\theta}^i$ for each input image $I^i$ adhere to the background knowledge $T$. Knowledge injection is performed via a logic-based loss function $\mathcal{L}^s$.
To scale to large theories, neural-guided projection (NGP) selects a fixed-size subset $T^{i^*}_{\rho}$ of the theory to compute the loss for each $I^i$.}
\label{Figure_overview}\label{fig:overview}
\end{figure} 

A \emph{scene graph} is a set of 
\textit{facts} describing the objects occurring in an image and their inter-relationships. \textit{Scene Graph Generation} (SGG) asks to identify all the facts that hold in an image.  
Using prior knowledge (for instance commonsense knowledge bases and knowledge graphs \cite{atomic}) is particularly appealing in SGG, as relationships in scene graphs naturally adhere to commonsense principles.
This intuition has led to the introduction of \emph{neurosymbolic} techniques \cite{DBLP:books/daglib/0007534} that inject background knowledge into a neural model at training-time and/or use it at inference-time  (also called \emph{testing-time}) to amend its predictions. 

Neurosymbolic SGG techniques are divided into two major families. The first one represents knowledge in a sub-symbolic fashion 
and incorporates it either only at training-time \cite{LENSR}, at testing-time \cite{Zareian-ECCV-2020}, or both at training- and testing-time \cite{DBLP:conf/cvpr/GuZL0CL19,Zareian_2020_ECCV}. The second family maintains knowledge in symbolic form and injects it into the model at training-time only \cite{LTN,van-Krieken-semi-supervised}.
While they have led to promising results, both groups of techniques face several shortcomings. The first one requires introducing ad-hoc, more complex neural architectures, and accessing the background knowledge at inference-time, thus increasing the training or testing cost. 
More importantly, ad-hoc neural architectures make it difficult to take advantage of state-of-the-art, neural SGG models, such as VCTree \cite{VCTREE}.
The second family suffers from limited scalability with respect to the number of formulas considered, making them impractical in real-world scenarios. 

Our work introduces a neurosymbolic regularization technique in which symbolic background knowledge, also referred to as a \textit{theory}, is used as an additional supervision signal for a neural model (see Figure \ref{fig:overview}). Our objective is to amend the neural network when its predictions do not abide by the background knowledge. 
{The main difference between our proposal and prior art on neurosymbolic SGG is that, instead of providing examples of what the neural model should predict (as in \cite{DBLP:conf/cvpr/GuZL0CL19,Zareian_2020_ECCV,Zareian-ECCV-2020}), we provide examples of what the model should \emph{not} predict. This is achieved by enforcing negative \emph{integrity constraints (ICs)}, expressed in the form 
$\neg \texttt{predicate}(\texttt{subject},\texttt{object})$, through a logic-based loss function.} 
The class of \emph{negative} ICs, which is not supported by 
\cite{DBLP:conf/cvpr/GuZL0CL19,Zareian_2020_ECCV,Zareian-ECCV-2020}
provides two benefits. 
Firstly, unlike any other symbolic-based regularization method,
it allows us to design a technique that scales in the presence of hundreds of thousands of ICs.
To this extent, instead of using the whole theory for regularizing every training sample, we propose a \emph{neural-guided projection (NGP)} procedure that identifies a small subset of ICs which are maximally logically violated under the neural predictions. The task of amending the neural module towards having its outputs abide by the ICs amounts to solving an optimization problem in which the weights of the neural module are updated to minimize the maximum violation of the ICs. 
{Secondly, it is easy for users to (semi-)automatically create such ICs from existing knowledge bases or even from the training data itself,
{by creating a negative IC out of each fact \textit{not} in the knowledge base or training data.}
To assess the robustness of NGP, we ran experiments using two different theories. The first one was created by taking the complement of the commonsense knowledge graph ConceptNet \cite{conceptnet}, while the second one  by taking the complement of the training facts.} 

Beyond outperforming prior relevant (sub)symbolic regularization techniques, NGP offers multiple other benefits. 
Firstly, unlike \cite{DBLP:conf/cvpr/GuZL0CL19,Zareian_2020_ECCV}, NGP is oblivious to the neural models and loss function {used}.
Furthermore, it does not require accessing the background knowledge at inference-time like \cite{DBLP:conf/cvpr/GuZL0CL19,Zareian_2020_ECCV}.  
Similarly to \cite{LENSR,LTN,DBLP:conf/cvpr/GuZL0CL19,Zareian_2020_ECCV,DBLP:conf/eccv/ZhuFF14}, as well as to prior art on knowledge distillation \cite{distillation2,distillation3}, we do \emph{not} question the background knowledge. Our analysis shows that NGP is robust to the theory in use, improving accuracy even when considering \emph{only} the complement of the training facts as negative ICs. Our empirical comparison confirms that NGP:
\begin{compactitem}
    \item improves the accuracy of state-of-the-art SGG models, namely IMP \cite{IMP}, MOTIFS \cite{MOTIFS} and VCTree \cite{VCTREE}, by up to $33\%$; 
    
    \item scales to theories including approximately ~1M ICs-- sizes {no prior symbolic-based regularization technique supports} \cite{LTN};
    
    \item is particularly effective when applied in conjunction with TDE \cite{TDE}, a technique that tackles the bias in the data, improving the performance of IMP, MOTIFS and VCTree by up to 16 percentile units;
    
    \item outperforms GLAT \cite{Zareian-ECCV-2020} and LENSR \cite{LENSR}, two state-of-the-art regularization techniques that maintain the knowledge in subsymbolic form, by up to $18\%$ and $15\%$;
    
    \item improves the accuracy of SGG models by up to six times when restricting the availability of ground-truth facts.
\end{compactitem}
Via {suitable} regularization components, such as TDE \cite{TDE}, we outperform in accuracy recently introduced state-of-the-art models \cite{BGNN} by up to $90\%$ and 
ad-hoc neurosymbolic SGG architectures leveraging external knowledge bases \cite{DBLP:conf/cvpr/GuZL0CL19} by up to $86\%$. 
\section{Preliminaries}

First-order logic is a language of \emph{predicates}, \emph{variables} and \emph{constants}.
\textit{Terms} are either variables or constants. 
An \emph{atom} $\alpha$ is an expression of the form $p(\vec{t})$, where $p$ is a predicate and $\vec{t}$ is a vector of terms.
\emph{Formulas} are expressions composed over atoms and the logical connectives, $\wedge$, $\vee$ and $\neg$; a formula is propositional if instead of atoms, it is composed over terms. 
A formula is \textit{ground} when it includes exclusively constants.  
We use ${t \in \varphi}$ to denote that a variable $t$ occurs in a propositional formula $\varphi$. 
A \emph{theory} $T$ is a set of formulas. The set of all possible atoms formed using the predicates
and the constants occurring in $T$ is the \emph{universe} $U$ of $T$. 
An \emph{interpretation} $J$ of $T$ is a total mapping from the elements in $U$ to a domain. We denote by $J(\varphi)$ the value of $\varphi$ in $J$.

{\textbf{Classical semantics}}
Interpretations $J$ in classical Boolean logic map elements in the universe to either true ($\top$) or false ($\bot$).
We say that $J$ \textit{satisfies} $\varphi$ if 
$\varphi$ evaluates to true in $J$, i.e., $J(\varphi)=\top$, and refer to $J$ as a \textit{model} of $\varphi$.

{\textbf{Fuzzy logic semantics}}
Interpretations in fuzzy logic map elements in the universe to the interval ${[0,1]}$. 
There are multiple ways\footnote{The truth of ground formula $\varphi$ is: ${J(\neg \varphi) \defeq 1 - J(\varphi)}$, ${J(\varphi_1 \wedge \varphi_2) \defeq \max\{0, J(\varphi_1) + J(\varphi_2) -1\}}$, ${J(\varphi_1 \vee \varphi_2) \defeq \min\{1, J(\varphi_1) + J(\varphi_2)\}}$ in Lukasiewicz t-(co)norms.} to define the logical connectives (see \cite{KR2020-92}).
We say that $J$ \textit{satisfies} $\varphi$ if $J(\varphi) = 1$. 

\textbf{Probabilistic semantics}
In probabilistic logics, similarly to the classical case, statements are either true or false. However, a probability is assigned to these truth values \cite{fuzzy-probability}. 
Consider a propositional formula $\varphi$ composed over independent Bernoulli random variables, where each variable $t$ is true with probability $p(t)$ and false with probability $1-p(t)$. 
Let $\mathbf{p}$ denote the vector of the probabilities so assigned to the variables.
The probability $P(J,\mathbf{p})$, of an interpretation $J$ under $\mathbf{p}$ is zero if $J$ is \emph{not} a model of $\varphi$; otherwise it is given by:
\begin{align} \label{eq:weight_interpretation}
    \prod_{\texttt{t} \in \varphi \mid J(\texttt{t}) = \top} {p}(\texttt{t}) \cdot \prod_{\texttt{t} \in \varphi \mid J(\texttt{t}) = \bot} 1 - {p}(\texttt{t})\;.
\end{align}  

Given \eqref{eq:weight_interpretation}, the probability of formula $\varphi$ being true under $\mathbf{p}$, denoted as ${P(\varphi|\mathbf{p})}$, is the sum of the probabilities of all the models of $\varphi$ under $\mathbf{p}$ (\cite{wmc}):
\begin{align} \label{eq:wmc}
P(\varphi|\mathbf{p}) =
    \sum \limits_{J \text{ model of } \varphi}\, P(J,\mathbf{p})\;.
\end{align}

\begin{example}
    Consider the formula ${\phi = \neg (\texttt{h} \wedge \texttt{d} \wedge \texttt{e})}$, 
    where \texttt{h} stands for horse, \texttt{d} for drinks and \texttt{e} for eye.  
    All interpretations of $\phi$,
    apart from the one assigning true to each variable, are models of the formula, i.e., the formula evaluates to true in those interpretations. 
    Assuming that each one of the above terms is assigned a probability $p(\cdot)$,
    the probability of the interpretation that assigns each variable to false is computed as
    ${(1 - {p}(\texttt{e})) \times (1 - {p}(\texttt{d})) \times (1 - {p}(\texttt{h}))}$. 
\end{example}
\section{Proposed framework}\label{section:framework}

Scene graph generation aims to identify all the \texttt{predicate}(\texttt{subject},\texttt{object}) facts that hold in an image.  
Let $\texttt{S}$, $\texttt{P}$ and $\texttt{O}$ be the sets of possible subject, predicate and object {terms}, respectively. Let also $n$ be a neural module that takes an input image and outputs the facts that are predicted to hold in that image. 
Without loss of generality, we assume that the output neurons of $n$ are divided into three mutually disjoint sets so that there is a one-to-one mapping between the neurons within each set and the elements included in sets $\texttt{S}$, $\texttt{P}$ and $\texttt{O}$. 
We use $\texttt{S}$, $\texttt{P}$ and $\texttt{O}$ to denote both the sets of terms and the sets of neurons mapped to those terms and use 
\texttt{t} to refer both to a term and to the neuron that maps to \texttt{t}. 
We denote by $w_{\theta}({\texttt{t}})$ the activation value of output neuron ${\texttt{t}}$, where $\theta$ denotes the trainable parameters of $n$,
and by $\mathbf{w}_{\theta}$ the vector of activation values of the output neurons, i.e., the predictions of $n$.

Facts in a scene graph usually abide by commonsense knowledge. 
We focus on commonsense knowledge encoded as a theory $T$ in first-order logic and in particular on theories in the form of \emph{integrity constraints} (ICs). 
Namely, an example of a \textit{negative} IC is the formula $\varphi$ given by ${\neg \texttt{drinks}(\texttt{horse},\texttt{eye})}$,
which expresses the restriction that a horse cannot drink an eye. 
Hereafter, we will consider $T$ to include exclusively \textit{negative}, \textit{atomic} ICs. 

{\textbf{Semantics}}
A theory $T$ can be used to penalize a model $n$.
For instance, penalizing $n$ under $\varphi$ involves adjusting $n$'s weights $\theta$ so that the neurons ${\texttt{drinks}}$, 
${\texttt{horse}}$ and ${\texttt{eye}}$ cannot \textit{simultaneously} take high activation values. 
In the language of logic, the terms in \texttt{S}, \texttt{P} and \texttt{O} form a {universe}. When adopting a probabilistic logic semantics, the activation values $\mathbf{w}_{\theta}$ of the output neurons can be seen as the likelihood $\mathbf{p} = \mathbf{w}_{\theta}$ of those terms. When adopting the semantics of fuzzy logic, instead, the vector $\mathbf{w}_{\theta}$ can be seen as an {interpretation} $J$ of the output terms as activation values map terms to the interval $[0,1]$,  see above.

\subsection{Loss functions} \label{subsection:loss-functions}

To inject background knowledge into a neural model, we need to quantify the level to which an IC $\varphi$ is \emph{consistent} with the neural predictions $\mathbf{w}_{\theta}$. In the case of probabilistic logic, we denote
this level of consistency by
$P(\varphi|\mathbf{w}_{\theta})$ (see \eqref{eq:wmc}). In fuzzy logic, 
we denote this level of consistency by $\mathbf{w}_{\theta}(\varphi)$, 
as $\mathbf{w}_{\theta}$ is treated as an {interpretation}.
Our framework is not bound to a specific semantics for interpreting theory $T$, adopting any semantics. To transparently support semantics that blend classical logic with uncertainty, we assume the existence of a function ${SAT: (\varphi,\mathbf{w}_{\theta}) \rightarrow R^+}$ expressing the degree of consistency of $\varphi$ with $\mathbf{w}_{\theta}$. 

Quantifying the consistency between $\varphi$ and $\mathbf{w}_{\theta}$ allows us to define a loss function ${\mathcal{L}^s(\varphi,\mathbf{w}_{\theta})}$ that is inversely proportional to $SAT(\varphi,\mathbf{w}_{\theta})$. 
Again, we do not stick to a specific loss function or semantics as in prior art, e.g., \cite{LTN}, but rather spell out the properties a loss function should satisfy to be incorporated into our framework: (i) ${\mathcal{L}^s(\varphi,\mathbf{w}_{\theta}) = 0}$ if the probability of 
$\varphi$ under $\mathbf{w}_{\theta}$ is one (in the case of probabilistic logic) or
$\mathbf{w}_{\theta}(\varphi) = 1$ (in the case of fuzzy logic); (ii) $\mathcal{L}^s$ is differentiable almost everywhere. 
The first property is to ensure the soundness of the loss function w.r.t. the logic semantics, while the second one is to ensure the ability to train via backprobagation.
We use $\mathcal{L}^s(T,\mathbf{w}_{\theta})$ as a shorthand for $\mathcal{L}^s(\bigwedge_{\varphi \in T} \varphi,\mathbf{w}_{\theta})$. 

\section{Considered Loss functions}\label{appendix:loss-functions}

\paragraph{(i) DL2 (Fuzzy logic semantics)}We considered the recently introduced fuzzy logic-based loss DL2 \cite{dl2}. The loss is differentiable almost everywhere and its gradients are more effective (i.e., non-zero) than those computed under other fuzzy logics, e.g., PSL \cite{psl-long}.
Below, we recapitulate the definition of DL2.

\begin{definition}[Adapted from \cite{dl2}] \label{definition:dl2}
Let $\mathcal{D}$ be a set of Boolean variables,
$X$ be a variable in $\mathcal{D}$, $\varphi$, $\varphi_1$ 
and $\varphi_2$ be formulas over variables in $\mathcal{D}$ and the Boolean connectives $\wedge$, $\vee$ and $\neg$, and $\psi$ be the formula that results after applying the De Morgan's rule 
to formula $\neg \varphi$ until negations are applied on the level of variables. 
Let also $\mathbf{w}$ be a vector assigning to each variable in  
$\mathcal{D}$ a value in ${[0,1]}$. 
$\mathcal{L}^s$ is defined as follows:
    \begin{align}
        \mathcal{L}^s(X,\mathbf{w}) &\defeq 1 - \mathbf{w}(X) \\
        \mathcal{L}^s(\neg X,\mathbf{w}) &\defeq \mathbf{w}(X) \label{eq:dl2:negation} \\
        \mathcal{L}^s(\varphi_1 \wedge \varphi_2,\mathbf{w}) &\defeq \mathcal{L}^s(\varphi_1,\mathbf{w}) + \mathcal{L}^s(\varphi_2,\mathbf{w}) \label{eq:dl2:conjunction}\\
        \mathcal{L}^s(\varphi_1 \vee \varphi_2,\mathbf{w}) &\defeq \mathcal{L}^s(\varphi_1,\mathbf{w}) \cdot \mathcal{L}^s(\varphi_2,\mathbf{w}) \label{eq:dl2:disjunction} \\
        \mathcal{L}^s(\neg \varphi,\mathbf{w}) &\defeq \mathcal{L}^s(\psi,\mathbf{w}) 
    \end{align}
\end{definition}

\paragraph{(ii) SL (Probabilistic logic semantics)}
To define a loss based on \eqref{eq:wmc}, we can 
employ standard cross entropy (as in \cite{tsamoura2020neuralsymbolic}). The cross entropy of \eqref{eq:wmc} is also known as \textit{semantic loss} (SL) \cite{semantic-loss}. An example of the computation of SL is shown in Table~\ref{table:running:wmc}. 

\begin{table}[h]
\caption{Computing the probability of formula ${\varphi = \neg (\texttt{eye} \wedge \texttt{drinks} \wedge \texttt{horse})}$ for a vector of neural predictions $\mathbf{w}$. \texttt{e} is short for \texttt{eye}, \texttt{d} is short for \texttt{drinks} and \texttt{h} is short for \texttt{horse}. $J$ denotes a Boolean interpretation of $\varphi$.} 
\label{table:running:wmc}
\scriptsize
\centering
	\begin{tabular}{ |c c c c | }
		\hline
		\texttt{eye}    & \texttt{drinks}       & \texttt{horse}     & ${P(J,\mathbf{w})}$ \\
		\hline
		$\bot$          & $\bot$       & $\bot$    &  ${(1 - {w}(\texttt{e})) \times (1 - {w}(\texttt{d})) \times (1 - {w}(\texttt{h}))}$ \\
        $\bot$          & $\bot$       & $\top$    &  ${(1 - {w}(\texttt{e})) \times (1 - {w}(\texttt{d})) \times {w}(\texttt{h})}$ \\
        $\bot$          & $\top$       & $\bot$    &  ${(1 - {w}(\texttt{e})) \times {w}(\texttt{d}) \times (1 - {w}(\texttt{h}))}$ \\
        $\bot$          & $\top$       & $\top$    &  ${(1 - {w}(\texttt{e})) \times {w}(\texttt{d}) \times {w}(\texttt{h})}$ \\
		$\top$          & $\bot$       & $\bot$    &  ${{w}(\texttt{e}) \times (1 - {w}(\texttt{d})) \times (1 - {w}(\texttt{h}))}$ \\
        $\top$          & $\bot$       & $\top$    &  ${{w}(\texttt{e}) \times (1 - {w}(\texttt{d})) \times {w}(\texttt{h})}$ \\
        $\top$          & $\top$       & $\bot$    &  ${{w}(\texttt{e}) \times {w}(\texttt{d}) \times (1 - {w}(\texttt{h}))}$ \\
        $\top$          & $\top$       & $\top$    &  0 \\
		\hline
	\end{tabular}
\end{table}

\subsection{Properties of loss functions} 

We summarize some properties for SL \cite{semantic-loss} and DL2 \cite{dl2}. Below, $\phi$, $\phi_1$, $\phi_2$ and $\mathbf{w}$ are as in Definition~\ref{definition:dl2}.

\begin{proposition}[From \cite{semantic-loss}]\label{proposition:SL}
SL satisfies the following properties:
\begin{itemize} 

	\item[] $\mathsf{P}_1$. SL is differentiable almost everywhere; 

    \item[] $\mathsf{P}_2$. ${\text{SL}(\varphi,\mathbf{w}) = 0}$, if $P(\varphi,\mathbf{w}) =1$; 
    
    \item[] $\mathsf{P}_3$. ${\text{SL}(\varphi_1,\mathbf{w}) \leq \text{SL}(\varphi_2,\mathbf{w})}$, if $P(\varphi_1 | \mathbf{w}) \geq P(\varphi_2 | \mathbf{w})$;
    
    \item[] $\mathsf{P}_4$. ${\text{SL}(\varphi_1,\mathbf{w}) = \text{SL}(\varphi_2,\mathbf{w})}$, if $\varphi_1$ is logically equivalent to $\varphi_2$;
\end{itemize}
\end{proposition}

\begin{proposition}[From \cite{dl2}]
DL2 satisfies the following properties:
\begin{itemize} 

	
	\item[] $\mathsf{P}_1$. DL2 is differentiable almost everywhere; 

    \item[] $\mathsf{P}_2$. ${\text{DL2}(\varphi,\mathbf{w}) = 0}$, if $\mathbf{w}(\varphi) =1$; 
    
    \item[] $\mathsf{P}_3$. ${\text{DL2}(\varphi_1,\mathbf{w}) \leq \text{DL2}(\varphi_2,\mathbf{w})}$, if $\mathbf{w}(\varphi_1) \geq \mathbf{w}(\varphi_2)$;
\end{itemize}
\end{proposition}
While SL satisfies $\mathsf{P}_4$ (this is due to the fact that $\varphi_1$ and $\varphi_2$ share the same models as they are logically equivalent) DL2 does not satisfy it. We provide an example demonstrating this case.
Consider the Boolean formula $\phi_1 = X_1 \wedge X_2 \vee X_1$. Formula $\phi_1$ is logically equivalent to the formula $\phi_2 = X_1$ as they both become true whenever $X_1$ is true. Despite that the probabilities, and consequently the probability-based losses, of both formulas will be the same as they share the same models, the losses computed according to DL2 will be different. 
The DL2 loss of formula $\phi_1$ is computed as 
$((1-{w}({X_1})) + (1-{w}({X_2})) ) \cdot (1-{w}({X_1}))$, while 
the DL2 loss of $\phi_2$  is computed as $1-{w}({X_1})$.

\subsection{Optimization objective}

We are now ready to introduce our technique. 
Let ${I^1,\dots,I^m}$ be a sequence of training images. SGG benchmarks such as Visual Genome (VG) \cite{VG} include for each image $I^i$ a ground truth set $\mathcal{F}^i$ of \texttt{predicate}(\texttt{subject},\texttt{object}) facts representing relationships that hold in $I^i$. State-of-the-art neural modules are trained based on loss functions $\mathcal{L}^n$ that take as arguments the facts in $\mathcal{F}^i$ and the neural predictions for $I^i$. We denote by $\mathbf{w}_{\theta}^i$ the predictions of $n$ for $I^i$. As
increasing the level of consistency between the ICs in $T$ 
and $\mathbf{w}_{\theta}^i$ 
reduces to minimizing the loss function $\mathcal{L}^s$, our optimization objective becomes:  
\begin{align}
	\theta^* \defeq arg\min \limits_{\theta}  \beta_1 \cdot \sum \limits_{i=1}^m \mathcal{L}^n(\mathcal{F}^i, \mathbf{w}^i_{\theta})   +  \beta_2 \cdot \sum \limits_{i=1}^m \mathcal{L}^s(T, \mathbf{w}^i_{\theta}). \nonumber
\end{align}
Above, $\beta_1$ and $\beta_2$ are hyperparameters setting the importance of each component of the loss.
In our empirical evaluation, those hyperparameters are computed in an automated fashion using \cite{weighting}. 
The loss function can be an arbitrary, non-linear function and hence  
${\mathcal{L}^s(T, \mathbf{w}^i_{\theta})}$ is not necessarily equal to 
${ \sum \limits_{\varphi \in T} \mathcal{L}^s(\varphi, \mathbf{w}^i_{\theta}) }$. 

\subsection{Neural-Guided projection}\label{sec:ngp}

Commonsense knowledge bases can be quite large. Hence, if naively implemented, regularization would be very time consuming if not infeasible. To overcome this limitation in a way that aligns with our optimization objective, for each training image $I^i$ we identify the subset $T^{i^*}_{\rho}$ of $\rho$ integrity constraints associated with the highest value of $\mathcal{L}^s$ among all possible subsets $T^{i}_{\rho}$ of $\rho$ ICs. 
We call the elements of $T^{i^*}_{\rho}$ the \textit{maximally non-satisfied ICs}:
\begin{align}
	T^{i^*}_{\rho} &\defeq arg\max_{T^i_{\rho} \subseteq T} \mathcal{L}^s(T^i_{\rho}, \mathbf{w}^i_{\theta}), \label{eq:constraints}
\end{align}
and regularize the neural module w.r.t. those constraints.
Regularizing using the maximally non-satisfied ICs maximizes our chances of providing meaningful feedback to the model. Consider again the IC ${\varphi = \neg \texttt{drinks}(\texttt{horse}, \texttt{eye})}$. 
If the likelihood of $\varphi$ being true under $\mathbf{w} _{\theta}$ is close to zero, then we are confident that the prediction needs to be amended; otherwise, we cannot know whether the neural predictions are indeed the correct ones or not and hence, we cannot provide meaningful feedback. 
In that case, only the ground truth annotations can provide meaningful supervision signal to the neural model.

Our technique, referred to as \emph{neural-guided projection (NGP)}, is summarized in Algorithm~\ref{algorithm:sgg_train}, which
presents the steps taking place on an image-by-image basis.
The algorithm denotes by $I$ the input image, 
by $\mathcal{F}$ the ground truth facts that hold in $I$, by $T$ the theory,
and by $n_t$ the state of the neural module at the $t$-round of the training process,
while $\rho$ defines the number of ICs to choose.
An overview of NGP is shown in Figure~\ref{fig:overview}.

\begin{algorithm}[tb] 
\caption{\textsc{NGP}($I, \mathcal{F}, T, n_t$) $\rightarrow$ $n_{t+1}$}  \label{algorithm:sgg_train}
\begin{algorithmic}[1]

    \State ${\mathbf{w} \defeq n_t(I)}$ 
    
    \State $T^*_{\rho} \defeq arg\max\limits_{T_{\rho} \subseteq T} \mathcal{L}^s(T_{\rho}, \mathbf{w})$
    
    \State $\ell \defeq \beta_1 \cdot \mathcal{L}^n(\mathcal{F}, \mathbf{w})   +  \beta_2 \cdot \mathcal{L}^s(T^*_{\rho}, \mathbf{w}) $
    
    \State $n_{t+1} \defeq \mathsf{backpropagate}(n_t, \bigtriangledown \ell)$
    
    \State \textbf{return} $n_{t+1}$
\end{algorithmic}
\hrule
\textbf{Note:} $\beta_1$, $\beta_2$ and $\rho$ are hyperparameters.
\end{algorithm} 

\textbf{Computing ${T^{i^*}_{\rho}}$} 
A greedy strategy for computing the set of maximally non-satisfied ICs is presented in Algorithm~\ref{algorithm:greedy}.
The arguments are as in Algorithm~\ref{algorithm:sgg_train}. 
Iteratively sampling $\rho$ constraints from the theory and computing $\mathcal{L}^s$ after taking the conjunction of those constraints is also an option. 

\begin{algorithm}[tb]
\caption{\textsc{GREEDY}($I, \rho, T, n_t$) $\rightarrow$ $T^*$} \label{algorithm:greedy}
\begin{algorithmic}[1]

    \State ${\mathbf{w} \defeq n_t(I)}$\; $T^* \defeq \emptyset$\; ${j \defeq 1}$  
    \While{$|T^*| < \rho$}
        \State \textbf{get} the $j$-th \texttt{p}(\texttt{s},\texttt{o}) prediction maximizing 
        $w(\texttt{p}) \cdot w(\texttt{s}) \cdot w(\texttt{o})$ \label{algorithm:greedy:selection}
        
        \State \textbf{if} {$\neg \texttt{p}(\texttt{s},\texttt{o})$ is in $T$}, \textbf{then} \textbf{add} $\neg \texttt{p}(\texttt{s},\texttt{o})$ to $T^*$ 
        \State $j \defeq j + 1$
    \EndWhile
    \State \textbf{return} $T^*$
\end{algorithmic}
\end{algorithm}

Proposition~\ref{proposition:maxT}
summarizes the cases in which the ICs chosen in Algorithm~\ref{algorithm:greedy} are the ones maximizing \eqref{eq:constraints}. Let $T^*$ be the set of ICs returned by Algorithm~\ref{algorithm:greedy}, SL denote the semantic loss and DL2 the fuzzy loss from \cite{dl2}.

\begin{proposition} \label{proposition:maxT}
	When $\mathcal{L}^s$=SL, then $T^*$ maximizes \eqref{eq:constraints} when  
	the formulas in $T^*$ share no common variables.
	When $\mathcal{L}^s$=DL2, then $T^*$ always maximizes \eqref{eq:constraints}.  
\end{proposition}

The proof is based on the notion of linearly-separable logic-based loss functions.
    \begin{definition} \label{definition:separability}
        Let $\phi_1$, $\phi_2$ and $\mathbf{w}$ be as in Definition~\ref{definition:dl2}.
    	A loss function $\mathcal{L}^s$ is linearly-separable if-f  $\mathcal{L}^s(\varphi_1 \wedge \varphi_2, \mathbf{w}) = \mathcal{L}^s(\varphi_1, \mathbf{w}) + \mathcal{L}^s(\varphi_2, \mathbf{w})$.
    \end{definition}
    When $\mathcal{L}^s$ is linearly-separable, then for a theory of closed formulas $T$, we have
    \begin{align}\mathcal{L}^s(\bigwedge \limits_{\varphi \in T}\varphi, \mathbf{w}) =  
    \sum \limits_{\varphi \in T} \mathcal{L}^s(\varphi, \mathbf{w})
    \end{align}
Furthermore, we will make use of the following result: 
    \begin{proposition} \label{proposition:linearly-separable}[From \cite{semantic-loss}]
    	Continuing with Definition~\ref{definition:separability}, SL is linearly-separable if-f $\varphi_1$ and $\varphi_2$ share no common variables.
    \end{proposition}
We are now ready to return to the main body of the proof. 

\begin{proof}
We distinguish the following cases:

\textbf{Case 1}.
    $\mathcal{L}^s$ is DL2. 
    Firstly, DL2 is linearly-separable due to \eqref{eq:dl2:conjunction}.
    Furthermore, from \eqref{eq:dl2:negation} and \eqref{eq:dl2:disjunction}, 
    it follows that the loss of an IC 
    $\neg(\texttt{p} \wedge \texttt{s} \wedge \texttt{o}) \equiv 
    \neg \texttt{p} \vee \neg \texttt{s} \vee \neg \texttt{o}$, 
    under $\mathbf{w}$, will be computed by 
    ${w}(\texttt{p}) \cdot {w}(\texttt{s}) \cdot {w}(\texttt{o})$
    The above, along with the fact that Algorithm~\ref{algorithm:greedy} chooses the 
    \texttt{p}(\texttt{s},\texttt{o}) predictions maximizing  
    ${w}(\texttt{p}) \cdot {w}(\texttt{s}) \cdot {w}(\texttt{o})$
    for which $\neg \texttt{p}(\texttt{s},\texttt{o})$ is in $T$, proves Proposition 1. 
    
\textbf{Case 2}. $\mathcal{L}^s$ is SL. 
    From line \ref{algorithm:greedy:selection} in Algorithm~\ref{algorithm:greedy} and from 
    \eqref{eq:wmc}, it follows that for a given $\mathbf{w}$, 
    Algorithm~\ref{algorithm:greedy} chooses the $j$-th \texttt{p}(\texttt{s},\texttt{o}) fact  maximizing
    $P(\texttt{p} \wedge \texttt{s} \wedge \texttt{o}| \mathbf{w})$. 
    Since by definition $P(\top| \mathbf{w}) = 1$, it follows that 
    the chosen prediction minimizes 
    $P(\neg (\texttt{p} \wedge \texttt{s} \wedge \texttt{o})| \mathbf{w})$, while
    due to Property $\mathsf{P}_3$ from Proposition~\ref{proposition:SL}, 
    it follows that the $j$-th prediction maximizes 
    $\mathcal{L}^s(\neg (\texttt{p} \wedge \texttt{s} \wedge \texttt{o}), \mathbf{w})$. 
    The above, along with Proposition~\ref{proposition:linearly-separable}, show that Proposition 1 holds when the formulas in $T^*$ share no common variables.
\end{proof}

\subsection{An example of NGP}
We provide an example of NGP applied in conjunction with Algorithm \ref{algorithm:sgg_train} and \ref{algorithm:greedy} for
greedily computing $T^{i^*}_{\rho}$.

\begin{example}
    Consider an input image $I^i$ as in Figure~\ref{fig:overview}, depicting a person wearing a jacket and being in front of a horse. Assume that theory $T$ consists of the following ICs:
    \begin{itemize}
        \item $\neg \texttt{drinks}(\texttt{horse}, \texttt{eye})$
        \item $\neg \texttt{wearing}(\texttt{horse}, \texttt{person})$
        \item $\neg \texttt{of}(\texttt{tail}, \texttt{person})$
        \item $\neg \texttt{eats}(\texttt{person}, \texttt{jacket})$
        \item $\neg \texttt{made\_of}(\texttt{tail}, \texttt{horse})$
        \item $\neg \texttt{made\_of}(\texttt{person}, \texttt{jacket})$
        \item $\neg \texttt{of}(\texttt{hat}, \texttt{horse})$
    \end{itemize}
    
    Assume that the six most likely facts returned by the neural model are:
    \begin{itemize}
        \item $\texttt{of}(\texttt{tail}, \texttt{horse})$, with likelihood $0.34$.
        \item $\texttt{wearing}(\texttt{horse}, \texttt{person})$, with likelihood $0.27$.
        \item $\texttt{looking\_at}(\texttt{person}, \texttt{tail})$, with likelihood $0.26$.
        \item $\texttt{made\_of}(\texttt{person}, \texttt{jacket})$, with likelihood $0.24$.
        \item $\texttt{wearing}(\texttt{person}, \texttt{hat})$, with likelihood $0.19$.
        \item $\texttt{made\_of}(\texttt{tail}, \texttt{horse})$, with likelihood $0.19$.
    \end{itemize}
    Above, the likelihood of each fact is computed by multiplying the neural confidences $\mathbf{w}$ of its constituting subject, predicate, and object, respectively.
    From the above facts, the ones that not adhering to the theory are
    \begin{itemize}
        \item $\texttt{wearing}(\texttt{horse}, \texttt{person})$,
        
        \item $\texttt{made\_of}(\texttt{person}, \texttt{jacket})$,  
        
        \item $\texttt{made\_of}(\texttt{tail}, \texttt{horse})$.
    \end{itemize}

    Prior art (e.g., \cite{LTN}) computes a logic-based loss using all the ICs in the theory, leading to prohibitively expensive computations. Instead, NGP chooses a subset of the ICs, making the computation much more efficient.
    Assuming $\rho=2$ in Equation~\ref{eq:constraints} and using the greedy technique from Algorithm~\ref{algorithm:greedy}, NGP will compute a loss by taking only the $\rho=2$ maximally non-satisfied ICs, see Equation~\ref{eq:constraints}. In our example, $T^{i^*}_{\rho}$ consists of the following ICs:
    \begin{itemize}
        \item $\neg \texttt{wearing}(\texttt{horse}, \texttt{person})$
        \item $\neg \texttt{made\_of}(\texttt{person}, \texttt{jacket})$
    \end{itemize}
    The logic-based loss function $\mathcal{L}^s$ is computed using $T^{i^*}_{\rho}$ and is added to the supervised loss $\mathcal{L}^n$ to obtain the final loss used for training.
\end{example}
\section{Experiments} \label{section:experiments}

\textbf{Benchmarks} 
Following previous works, e.g., \cite{Zareian_2020_ECCV,BGNN}, we use Visual Genome (\textbf{VG}) \cite{VG} with the same split adopted by \cite{TDE}, and 
the Open Images v6 (\textbf{OIv6}) benchmark \cite{OIv4} with the same split adopted by \cite{BGNN}. 
Visual genome (VG) \cite{VG} includes 108K images across 75K object and 37K predicate categories. 
However, as 92$\%$ of the predicates have
no more than ten instances, we followed the widely adopted
VG split containing the most frequent 150
object categories and 50 predicate categories.
has in total 126K images for training and 1,813 and 5,322 images for validation and testing. In total, 301 object and 31 predicate categories are included.
In both benchmarks, each training, validation and testing datum is of the form ${(I,\mathcal{F})}$, where $I$ is an image and $\mathcal{F}$ is a set of \texttt{predicate}(\texttt{subject},\texttt{object}) facts relevant to $I$. The objects in each fact in $\mathcal{F}$ are annotated with 
their surrounding bounding boxes. 
We mostly focus on VG, as it is heavily biased \cite{TDE} and more challenging than OIv6 (SGG models have lower performance for VG than for OIv6, as also reported in \cite{BGNN}). 

{\textbf{Theories} 
We used $\mathbf{\theoryb}$ and $\mathbf{\theorya}$.
$\theoryb$ was computed by taking the complement of the training facts: 
we enumerated all combinations of predicates, subjects and objects in VG and for each \texttt{p}(\texttt{s},\texttt{o}) fact that is \emph{not} in the set of training facts, where \texttt{p}, \texttt{s} and \texttt{o} denotes a predicate, subject and object in the domain of VG, we added to $\theoryb$ the IC ${\neg \texttt{p}(\texttt{s},\texttt{o})}$.
We adopted the same approach to create theory $\theorya$ out of ConceptNet's knowledge graph. However, there we considered sparse subgraphs of the entire graph. In particular, 
we identified subject-object pairs (\texttt{s}, \texttt{o}) having less than ten \texttt{p}(\texttt{s}, \texttt{o}) facts in ConceptNet, where \texttt{p}, \texttt{s} and \texttt{o} denotes a predicate, subject and object in the domain of VG or OIv6, respectively. We then repeated the same process for subject-predicate and predicate-object pairs.}
While the presence, or absence, of a fact in either ConceptNet or the VG training data affects our theory, NGP is not biased by the training facts' {frequencies}.
We did not manually check the resulting theories and hence, they may include constraints that violate commonsense, reflecting real-world noisy settings. 
Theories $\theorya$ and $\theoryb$ include approximately 500k and 1M ICs.

\textbf{Models} 
Similarly to \cite{TDE} and \cite{EnergySGG}, we applied NGP on three state-of-the-art neural SGG models: \textbf{IMP} \cite{IMP}, \textbf{MOTIFS} \cite{MOTIFS} and \textbf{VCTree} \cite{VCTREE}. Prior art \cite{Zareian-ECCV-2020, TDE, BGNN} also considers KERN \cite{KERN} and VTransE \cite{VTranceE}-- we use the more recent model VCTree. 

\textbf{Regularization techniques} We considered several recently proposed state-of-the-art regularization techniques:
\begin{compactitem}
	\item \textbf{TDE} \cite{TDE}, a neural-based technique that operates at inference-time and aims at removing the bias towards more frequently appearing predicates in the data;
	\item \textbf{GLAT} \cite{Zareian-ECCV-2020}, a neural-based technique that amends SGG models at inference-time using patterns captured from the training facts;
	\item \textbf{LENSR} \cite{LENSR}, a neural-based technique that amends SGG models at training-time after embedding the input symbolic knowledge into a manifold;
	\item \textbf{LTNs} \cite{LTN}, a symbolic-based technique that injects the input symbolic knowledge to an SGG model at training-time;
	\item \textbf{$\customitr$}, our own symbolic-based technique that returns the most-likely prediction not violating any input IC, where the likelihood of a prediction is the product of the confidences of its predicate, subject and object as assigned by a model. $\customitr$ is an inference-time counterpart to NGP.
\end{compactitem}
LTNs is a direct competitor to NGP, while LENSR and GLAT are the neural counterpart to NGP.  
TDE does not use commonsense knowledge and hence it is orthogonal to all the other regularization techniques.

\textbf{Additional architectures} 
We consider \textbf{KBFN} \cite{DBLP:conf/cvpr/GuZL0CL19}, a state-of-the-art ad-hoc, architecture accessing ConceptNet both at training- and at testing-time; and \textbf{BGNN} \cite{BGNN} a recently-introduced confidence-aware bipartite graph neural network
with adaptive message propagation mechanism. 
In contrast to IMP, MOTIFS and VCTree, BGNN cannot be easily integrated with regularization techniques, as it makes use of an ad-hoc data sampling procedure at training-time. Indeed, the authors position BGNN as an alternative to models trained with TDE, and our empirical comparison manifests that it also does not integrate effectively with LENSR and GLAT.
 
\textbf{Computational environment}
All experiments ran on a Linux machine with 8 NVidia  GeForce GTX 1080 Ti GPUs, 64 Intel(R) Xeon(R) Gold 6130 CPUs, and 256GB of RAM.
 
\textbf{Overview of experimental results}
We considered the standard tasks of \emph{predicate} and \emph{scene graph classification}. Given an input image, and a set of bounding boxes with labels indicating the subjects/objects contained in each bounding box, predicate classification asks to predict the facts that hold in the image. In scene graph classification, the goal is the same, but the bounding boxes are unlabeled.
We used the standard measures {Mean Recall@$k$} (mR@$k$) and {zero-shot Recall@k} (zsR@$k$) to assess accuracy. mR@$k$ was proposed as a replacement to recall@$k$ to address the data bias issue in SGG benchmarks \cite{TDE,VCTREE}. 
zsR@k measures {recall@}$k$ considering only the facts that are in the testing but not the training set \cite{lu2016visual}.

We employed NGP with different loss functions. We set the number of constraints (i.e., $\rho$ in Eq. \ref{eq:constraints}) to $\rho=3$. We found that this value adds minimum computational overhead while improving mR and zsR. We considered the loss functions DL2 \cite{dl2} (fuzzy logic) and SL \cite{semantic-loss} (probabilistic logic). NGP(X) denotes NGP employed using loss X. All experiments ran using the full theories. LTNs were prohibitively slow for the size of our theory: using the same computational resources we used for NGP, it would have taken 4,000 hours for training for just one epoch. As such, we do not report results for LTNs. 

Table~\ref{tab:NGP-prob-vs-fuzzy-conceptnet} shows the impact of NGP, LENSR and ITR on IMP, MOTIFS and TDE for theory $\theorya$. Similarly, Table~\ref{tab:glat} shows the impact of NGP, GLAT and LENSR for theory $\theoryb$. 
NGP and LENSR adopt $\theoryb$ for a fair comparison against GLAT, as the latter regularizes SGG models using knowledge mined from the training images. 
Table~\ref{tab:NGP-tde-conceptnet} studies the integration of TDE and NGP on MOTIFS and VCTree (TDE does not support IMP~\cite{TDE}). 

The above results are on the VG dataset. Table~\ref{tab:NGP-TDE-motifs-oiv6} shows the impact of NGP(SL) with $\theorya$ and TDE on MOTIFS for the OIv6 dataset when the models are trained with limited access to the ground truth labels. In particular, we remove 0\%, 50\% and 75\% of the ground-truth facts at training-time, while keeping the corresponding images in the training set. As all the baselines we consider require the ground facts to compute a loss $\mathcal{L}^n$, the above setting leads to discarding each sample that misses ground-truth facts when training a baseline model (both with and without TDE). In contrast, when applying NGP, we use only $\mathcal{L}^s$ at training-time when the ground-truth facts are not available. The above setting demonstrates the effectiveness of NGP in weak supervision. We report results for MOTIFS, as it was the most challenging to regularize, as discussed below. 
OIv6 does not provide zero-shot evaluation and, thus, we report only mR@k.     
Similarly to Table~\ref{tab:NGP-TDE-motifs-oiv6}, Figure~\ref{fig:varying-sizes} shows the impact of NGP(SL) with $\theorya$ on IMP and VCTree when reducing 10\%--50\% of the ground-truth facts in VG. The task of interest is predicate classification. 
Again, when the ground-truth facts of an image are missing, $\mathcal{L}^s$ is used to back propagate through the SGG model when regularizing under NGP; images that miss ground-truth facts are ignored in the absence of NGP.
For completeness, Figure~\ref{fig:varying-sizes} also shows mR and zsR when using the full training set (0\% reduction).
Figure~\ref{fig:kbfn} reports results on VG for the ad-hoc architecture KBFN and the model BGNN. NGP is applied with $\theorya$ and KBFN with ConceptNet-- KBFN does not support negative ICs.

The appendix provides further details including implementation details, results on the integration of BGNN with LENSR and GLAT, and an analysis of the effects of different $\rho$'s.

\subsection{Key conclusions}

\begin{table*}[tb]
    \centering
    \caption{Impact of different regularization strategies on models' accuracy using $\theorya$. Results on the VG dataset.}
    \label{tab:NGP-prob-vs-fuzzy-conceptnet}
    \resizebox{0.75\textwidth}{!}
    {%
    \begin{tabular}{lllcccccc|cccccc}
        \toprule
        Model & Theory & Reg. & \multicolumn{6}{c}{Predicate Classification} & \multicolumn{6}{c}{Scene Graph Classification} \\
        & & & \multicolumn{3}{c}{mR@} & \multicolumn{3}{c}{zsR@} & \multicolumn{3}{c}{mR@} & \multicolumn{3}{c}{zsR@} \\
        & & & 20 & 50 & 100 & 20 & 50 & 100 & 20 & 50 & 100 & 20 & 50 & 100 \\
        \midrule
        IMP & - & - & 9.26 & 11.43 & 12.23 & 12.23 & 17.28 & 19.92 & 5.57 & 6.31 & 6.74 & 2.04	& 3.47 & 3.90 \\
        IMP & $\theorya$ & $\customitr$ & 9.27 & 11.44 & 12.23 & 12.24 & 17.30 & 19.94 & 5.61 & 6.35 & 6.78 & 2.08	& 3.50 & 3.92 \\
        IMP & $\theorya$ & LENSR & 10.56 & 13.16 & 14.22 & 12.78 & 18.31 & 21.06 & 0.01 & 0.01 & 0.02 & 0.01 & 0.01 & 0.01 \\
        IMP & $\theorya$  & NGP(SL) & {11.29} & {14.22} & {15.30} & {12.84} & \textbf{18.75} & {21.84} & \textbf{{6.99}} & \textbf{{8.45}} & \textbf{{8.92}} & \textbf{{2.71}} & \textbf{{4.48}} & \textbf{{5.35}} \\
        IMP & $\theorya$  & NGP(DL2) & \textbf{11.62} & \textbf{14.73} & \textbf{15.92} & \textbf{13.13} & {18.57} & \textbf{21.87} & 5.58 & 6.42 & 6.95 & 2.17 & 3.50 & 3.93 \\
        \midrule
        MOTIFS & & - & 12.65 & 16.08 & 17.35 & 1.21 & 3.34 & 5.57 & 6.81 & 8.31 & 8.85 & 0.33 & 0.65 & 1.13 \\
        MOTIFS & $\theorya$ & $\customitr$ & 12.68 & 16.10 & 17.39 & 1.23 & 3.35 & 5.57 & 6.82 & 8.32 & 8.85 & 0.35 & 0.66 & 1.13\\
        MOTIFS & $\theorya$ & LENSR & 12.50 & 15.90 & 17.20 & 1.12 & 3.26 & 5.37 & 0.30 & 0.34 & 0.36 & 0.02 & 0.02 & 0.02 \\ 
        MOTIFS & $\theorya$ & NGP(SL) & \textbf{{12.94}} & \textbf{{16.44}} & \textbf{{17.76}} & \textbf{{1.31}} & \textbf{{3.57}} & \textbf{{5.74}} & \textbf{{8.16}} & \textbf{{10.00}} & \textbf{{10.54}} & \textbf{{0.49}} & \textbf{{1.05}} & \textbf{{1.58}} \\
        MOTIFS & $\theorya$  & NGP(DL2) & 7.35 & 10.52 & 12.34 & 0.27 & 0.67 & 1.20 & 4.92 & 7.99 & 6.56 & 0.13 & 0.24 & 1.09 \\
        \midrule
        VCTree & - & - & 13.07 & 16.75 & 18.11 & 1.04 & 3.28 & 5.52 & 9.29 & 11.42 & 12.12 & 0.48 & 1.37 & 2.09 \\
        VCTree & $\theorya$ & $\customitr$ & 13.71 & 17.27 & 18.58 & \textbf{1.37} & 3.80 & \textbf{6.38} & 9.36 & 11.49 & 12.19 & 0.51 & 1.40 & 2.17\\
        VCTree & $\theorya$ & LENSR & 13.53 & 16.98 & 18.27 & {1.33} & 3.83 & 5.88 & 0.0 & 0.01 & 0.01 & 0.02 & 0.02 & 0.02 \\
        VCTree & $\theorya$ & NGP(SL) & {13.69} & \textbf{{17.51}} & \textbf{{18.92}} & {{1.29}} & \textbf{{3.85}} & {{6.04}} & \textbf{{9.89}} & \textbf{{11.75}} & \textbf{{12.35}} & \textbf{{0.67}} & \textbf{{1.56}} & \textbf{{2.44}}\\
        VCTree & $\theorya$  & NGP(DL2) & \textbf{13.86} & 17.49 & 18.77 & 1.16 & 3.62 & 5.68 & 9.41 & 11.56 & 12.12 & 0.49 & 1.38 & 2.39 \\
        \bottomrule
    \end{tabular}
    }
\end{table*}

\begin{table*}[t]
    \centering
    \caption{Impact of different regularization strategies on model's accuracy using $\theoryb$. Results on the VG dataset.}
    \label{tab:glat}
    \resizebox{0.75\textwidth}{!}{%
    \begin{tabular}{cllcccccc|cccccc}
        \toprule
        Model & Theory & Reg. & \multicolumn{6}{c}{Predicate Classification} & \multicolumn{6}{c}{Scene Graph Classification} \\
        & & & \multicolumn{3}{c}{mR@} & \multicolumn{3}{c}{zsR@} & \multicolumn{3}{c}{mR@} & \multicolumn{3}{c}{zsR@} \\
        & & & 20 & 50 & 100 & 20 & 50 & 100 & 20 & 50 & 100 & 20 & 50 & 100 \\
        \midrule
		IMP & - & - & 9.26 & 11.43 & 12.23 & 12.23 & 17.28 & 19.92 & 5.57 & 6.31 & 6.74 & 2.04	& 3.47 & 3.90 \\
		IMP & - & GLAT & 10.04 & 12.44 & 13.30 & 11.87 & 17.04 & 19.72 & 5.95 & 6.75 & 7.17 & 2.09 & 3.40 & 3.82\\
		IMP & $\theoryb$ & LENSR & 10.51 & 13.29 & 14.33 & 12.40 & 18.07 & 21.22 & 0.01 & 0.01 & 0.02 & 0.01 & 0.01 & 0.01\\
        		IMP & $\theoryb$  & NGP(SL) & \textbf{11.82}  & \textbf{15.16} & \textbf{16.46} & \textbf{12.39} & \textbf{18.18} & \textbf{{21.13}} & \textbf{7.14} & \textbf{8.60} & \textbf{9.15} & \textbf{2.95} & \textbf{4.62} & \textbf{5.66} \\
                  \midrule
	    	MOTIFS & - & - & 12.65 & 16.08 & 17.35 & 1.21 & 3.34 & 5.57 & 6.81 & 8.31 & 8.85 & 0.33 & 0.65 & 1.13 \\
		MOTIFS & - & GLAT & \textbf{12.82} & \textbf{16.26} & \textbf{17.60} & 1.26 & \textbf{3.49} & \textbf{5.79} & \textbf{6.84} & \textbf{8.34} & \textbf{8.89} & \textbf{0.32} & \textbf{0.63} & \textbf{1.12}\\
		MOTIFS & $\theoryb$ & LENSR & 12.57 & 16.09 & 17.38 & \textbf{1.37} & 3.41 & 5.65 & 0.01 & 0.01 & 0.01 & 0.02 & 0.02 & 0.02\\
        		MOTIFS & $\theoryb$  & NGP(SL) & 12.10 & 15.28 & 16.54 & {1.34} & 3.43 & 5.47 & 6.27 & 7.94 & 8.42 & 0.14 & 0.35 & 0.55 \\
	         \midrule
		VCTree & - & - & 13.07 & 16.75 & 18.11 & 1.04 & 3.28 & 5.52 & 9.29 & 11.42 & 12.12 & 0.48 & 1.37 & 2.09\\
		VCTree & - & GLAT & {13.88} & {17.51} & 18.90 & 1.28 & {3.87} & \textbf{6.43} & 9.39 & 11.52 & 12.20 & 0.51 & 1.42 & 2.17\\
		VCTree & $\theoryb$ & LENSR & 13.46 & 17.06 & 18.49 & 1.27 & 3.69 & 5.98 & 0.0 & 0.01 & 0.02 & 0.01 & 0.01 & 0.01\\
	        VCTree & $\theoryb$  & NGP(SL) & \textbf{14.09} & \textbf{17.72} & \textbf{19.08} & \textbf{1.35} & \textbf{3.98} & 6.36 & \textbf{9.57} & \textbf{11.68} & \textbf{12.49} & \textbf{0.61} & \textbf{1.51} & \textbf{2.39} \\
        \bottomrule
    \end{tabular}
    }
\end{table*}    

\conclude{NGP can substantially improve the recall of SGG models.} 
Table~\ref{tab:NGP-prob-vs-fuzzy-conceptnet} shows that NGP with theory $\theorya$ improves the relative mR@k of IMP, MOTIFS and VCTree up to $25\%$, $3\%$ and $4.5\%$ on predicate classification; on scene graph classification, the improvements are up to $33\%$, $20\%$ and $6.4\%$.
Table~\ref{tab:glat} shows that when NGP uses $\theoryb$, the relative improvements over IMP, and VCTree further increase to $34\%$ and $5\%$ on predicate classification, and to $36\%$ and $3\%$ on scene graph classification.
Table~\ref{tab:NGP-TDE-motifs-oiv6} shows that NGP can improve the performance of MOTIFS by~$4\%$ in predicate classification, even with fewer ground-truth facts. 
The results in Table~\ref{tab:NGP-prob-vs-fuzzy-conceptnet} for zsR@k also show that NGP can improve a model's generalization capabilities of predicting 
facts that are missing from the training set. 

We observe MOTIFS is sensitive to regularization:
LENSR always decreases its recall; NGP increases its recall with $\theorya$, but decreases it when adopting either $\theoryb$, see Table~\ref{tab:glat}, or the semantics of fuzzy logic, see Table~\ref{tab:NGP-prob-vs-fuzzy-conceptnet}. We conjecture that the decreases are because MOTIFS favors the most frequent predicate for a given subject-object pair in the ground-truth facts. Hence, adding a regularization term that penalizes predictions outside of the training facts may lead to severe overfitting explaining also the drastic drop in zsR@k. Regarding the fuzzy logic semantics, the decrease stresses the limitations of techniques like LTNs that are bound to fuzzy logic. Given the above, we only consider probabilistic logic for NGP hereafter, without discarding the potential of fuzzy logic in other scenarios.

\conclude{NGP outperforms prior regularization techniques in most scenarios.}
NGP is the most effective regularization technique in most cases in Table~\ref{tab:NGP-prob-vs-fuzzy-conceptnet}.  
For instance, regularization of IMP via NGP(SL) leads to up to $25\%$ higher mR@k over $\customitr$ on predicate classification. 
With the exception of MOTIFS, NGP also outperforms GLAT and LENSR in the scenarios in Table~\ref{tab:glat} leading to up to $20\%$ and $27\%$ higher accuracy in predicate and scene graph classification.
The results show that LENSR fails to provide a meaningful loss for training for scene graph classification. Below, we attempt to explain why. In advance of regularization, LENSR learns a manifold $\mathcal{M}$ representing the input theory and a function $q$ mapping embeddings of predictions into the space of $\mathcal{M}$. At regularization-time, LENSR maps via $q$ the embedding of a \texttt{p}(\texttt{s},\texttt{o}) prediction into the space of $\mathcal{M}$, where the embedding of \texttt{p}(\texttt{s},\texttt{o}) is the sum of the word embeddings of \texttt{s}, \texttt{p} and \texttt{o} weighted by $w(\texttt{p})$, $w(\texttt{s})$ and $w(\texttt{o})$. The L2 distance between the mapped embedding and $\mathcal{M}$ serves as a loss to back-propagate through an SGG model. As the predictions of a model have higher uncertainty in scene graph classification than in predicate classification (not only \texttt{p}, but also \texttt{s} and \texttt{o} are now uncertain), the embedding of \texttt{p}(\texttt{s},\texttt{o}) will be further away from the prediction embeddings that LENSR has seen while learning $q$ in advance of regularization. This discrepancy leads $q$ to transform the prediction embeddings erroneously, leading to a loss function that provides a meaningless training signal. 
LENSR was not tested on scene graph classification by the authors \cite{LENSR}.

\begin{table*}[tb]
    \small
    \centering
    \caption{Impact of NGP(SL) on MOTIFS and VCTree with TDE. Results on the VG dataset.}
    \label{tab:NGP-tde-conceptnet}
    \resizebox{0.75\textwidth}{!}{%
    \begin{tabular}{lllcccccc|cccccc}
        \toprule
        Model & Theory & Reg. & \multicolumn{6}{c}{Predicate Classification} & \multicolumn{6}{c}{Scene Graph Classification} \\ 
        & & & \multicolumn{3}{c}{mR@} & \multicolumn{3}{c}{zsR@} & \multicolumn{3}{c}{mR@} & \multicolumn{3}{c}{zsR@} \\
         & & & 20 & 50 & 100 & 20 & 50 & 100 & 20 & 50 & 100 & 20 & 50 & 100\\
        \midrule
        MOTIFS       & - & - & 12.65 & 16.08 & 17.35 & 1.21 & 3.34 & 5.57 & 6.81 & 8.31 & 8.85 & 0.33 & 0.65 & 1.13\\
        MOTIFS       & - & TDE & 17.18 & 23.95 & 27.66 & 8.10 & 13.68 & 17.11 & 10.11 & 13.44 & 15.35 & 1.85 & 3.01 & 3.68\\ 
        MOTIFS       & $\theorya$ & NGP(SL)+TDE & \textbf{17.99} & \textbf{24.50} & \textbf{28.16} & \textbf{8.51} & \textbf{14.00} & \textbf{17.80} & \textbf{11.80} & \textbf{15.11} & \textbf{16.77} & \textbf{1.92} & \textbf{3.05} & \textbf{3.74}\\
        \midrule
        VCTree       & - & - & 13.07 & 16.75 & 18.11 & 1.04 & 3.28 & 5.52 & 9.29 & 11.42 & 12.12 & 0.48 & 1.37 & 2.09\\
        VCTree       & - & TDE & 19.40 & 25.94 & 29.48 & 8.14 & 12.38 & 14.07 & 10.51 &  14.53 & 16.73 & 1.48 & 2.54 & \textbf{3.99} \\
        VCTree       & $\theorya$ & NGP(SL)+TDE & \textbf{23.91} & \textbf{30.78} & \textbf{34.19} & \textbf{8.15} & \textbf{12.47} & \textbf{15.41} & \textbf{13.60} & \textbf{17.69} & \textbf{19.85} & \textbf{1.57} & \textbf{2.63} & 3.63\\
        \bottomrule
    \end{tabular}
    }
\end{table*}

\conclude{NGP complements bias reduction techniques.}
Regarding MOTIFS, the recall improvements brought by TDE are up to $59\%$ and $73\%$ in predicate and scene graph classification and increase to $62\%$ and $89\%$ when NGP is additionally applied using $\theorya$.
Regarding VCTree, the recall improvements brought by TDE are up to $62\%$ and $38\%$ in predicate and scene graph classification; when NGP is additionally applied, recall increases up to $88\%$ and $63\%$. Tables~\ref{tab:NGP-prob-vs-fuzzy-conceptnet} and \ref{tab:NGP-tde-conceptnet} show that the combination of TDE with NGP leads to much higher improvements than the sum of the improvements obtained by applying each technique separately. 

\begin{table}[tb]
    \centering
    \caption{Impact of NGP(SL) and TDE on MOTIF when reducing the ground-truth facts from the OIv6 dataset.}
    \label{tab:NGP-TDE-motifs-oiv6}
    \resizebox{0.9\columnwidth}{!}
    {%
    \begin{tabular}{llccc|ccc}
        \toprule
        \% Red. & Reg. & \multicolumn{3}{c}{Prd Cls mR@} & \multicolumn{3}{c}{Sg Cls mR@}  \\
        & & 20 & 50 & 100 & 20 & 50 & 100 \\
        \midrule
        -0\% & - & 45.62 & 46.10 & 46.15 & \textbf{28.83} & \textbf{28.90} & \textbf{28.92}\\
        -0\% & TDE & 41.85 & 42.00 & 42.01 & 12.80 & 12.87 & 12.87 \\
        -0\% & NGP(SL) & \textbf{48.15} & \textbf{48.65} & \textbf{48.70} & 25.79 & 26.07 & 26.10 \\
        \midrule
        -50\% & - & 42.63  & 43.12 & 43.17 & 25.75  & 25.81  & 25.83 \\
        -50\% & TDE & 31.91 & 32.07 & 32.08 &  19.31 & 19.34 & 19.36  \\
        -50\% & NGP(SL) & \textbf{45.93} & \textbf{46.41}  & \textbf{46.46}  & \textbf{25.92}  & \textbf{26.21} & \textbf{26.24} \\
         \midrule
        -75\% & - & 41.97  & 42.16 & 42.17 & 23.21 & 23.27 & 23.28 \\
        -75\% & TDE  & 33.93 & 34.09 & 34.10 & 14.67  & 14.73 & 14.74 \\
        -75\% & NGP(SL) & \textbf{44.40} &  \textbf{44.90} &  \textbf{44.94} & \textbf{24.94} & \textbf{25.23} & \textbf{25.26} \\
        \bottomrule
    \end{tabular}
    }
\end{table}

\begin{figure}[tb]
\hspace{-0.3cm}
    \begin{adjustbox}{max width=1.05\columnwidth}
        \begin{tabular}{ccc}
             & {\huge \textbf{IMP}} & {\huge \textbf{VCTree}}\\
            \rotatebox{90}{\hspace{20mm}{\huge \textbf{Recall}}} &
            \x\x {\begin{tikzpicture}
\pgfplotstableread{
0 12.23
10 7.49
20 6.89
30 6.63
40 6.46
50 5.36
}\impmrhundred

\pgfplotstableread{
0 15.30
10 15.13
20 13.91
30 13.55
40 11.3
50 11.28
}\impmrhundredngp

\pgfplotstableread{
0 19.92
10 4.79
20 4.48
30 4.38
40 4.42
50 3.06
}\impzsrhundred

\pgfplotstableread{
0 21.84
10 21.01
20 21.34
30 21.03
40 18.53
50 18.99
}\impzsrhundredngp

\begin{axis}[
    font=\Large,
    xmin=-5, xmax=55,
    ymin=0, ymax=25,
    xtick={0,10,20,30,40,50},
    xticklabels={0\%,-10\%,-20\%,-30\%,-40\%,-50\%},   
    ytick={5,10,15,20},
    yticklabels={5\%,10\%,15\%,20\%},   
]

\addplot [
    color=blue, 
    mark=*,
    mark options={
    	draw=blue,
    	fill=blue
    },
    nodes near some coords={0/12.23/above,1/7.46/above,2/6.89/above,3/6.63/above,4/6.46/above,5/5.36/above}
] table {\impmrhundred};

\addplot [
    color=blue, 
    mark=*,
    dotted,
    mark options={
    	draw=blue,
    	fill=blue
    },
    nodes near some coords={0/15.30/above,1/15.13/above,2/13.91/above,3/13.55/above,4/11.3/above,5/11.28/above}
] table {\impmrhundredngp};

\addplot [
    color=orange, 
    mark=*,
    mark options={
    	draw=orange,
    	fill=orange
    },
    nodes near some coords={0/19.92/below,1/4.79/below,2/4.48/below,3/4.38/below,4/4.42/below,5/3.06/below}
] table {\impzsrhundred};

\addplot [
    color=orange, 
    mark=*,
    dotted,
    mark options={
    	draw=orange,
    	fill=orange
    },
    nodes near some coords={0/21.84/above,1/21.01/above,2/21.34/above,3/21.03/above,4/18.53/above,5/18.99/above}
] table {\impzsrhundredngp};

\end{axis}
\end{tikzpicture}} &
            \x\x {\begin{tikzpicture}
\pgfplotstableread{
0  18.11
10 17.03
20 14.54
30 15.01
40 14.55
50 13.39
}\vctreemrhundred

\pgfplotstableread{
0 18.92
10  17.96
20 16.45
30 16.96
40 16.16
50 14.86
}\vctreemrhundredngp

\pgfplotstableread{
0 5.52
10 5.83
20 4.79
30 4.65
40 4.34
50 4.72
}\vctreezsrhundred

\pgfplotstableread{
0 6.04
10  5.73
20 5.57
30 5.49
40 5.38
50 4.83
}\vctreezsrhundredngp

\begin{axis}[
    font=\Large,
    xmin=-5, xmax=55,
    ymin=2, ymax=21,
    xtick={0,10,20,30,40,50},
    xticklabels={0\%,-10\%,-20\%,-30\%,-40\%,-50\%},   
    ytick={5,10,15,20},
    yticklabels={5\%,10\%,15\%,20\%},   
]

\addplot [
    color=blue, 
    mark=*,
    mark options={
    	draw=blue,
    	fill=blue
    },
    nodes near some coords={0/18.11/below,1/17.03/below,2/14.54/below,3/15.01/below,4/14.55/below,5/13.39/below}
] table {\vctreemrhundred};

\addplot [
    color=blue, 
    mark=*,
    dashed,
    mark options={
    	draw=blue,
    	fill=blue
    },
    nodes near some coords={0/18.92/above,1/17.96/above,2/16.54/above,3/16.96/above,4/16.16/above,5/14.86/above}
] table {\vctreemrhundredngp};

\addplot [
    color=orange, 
    mark=*,
    mark options={
    	draw=orange,
    	fill=orange
    },
    nodes near some coords={0/5.52/below,1/5.83/below,2/4.79/below,3/4.65/below,4/4.34/below,5/4.72/below}
] table {\vctreezsrhundred};

\addplot [
    color=orange, 
    mark=*,
    dashed,
    mark options={
    	draw=orange,
    	fill=orange
    },
    nodes near some coords={0/6.04/above,1/5.73/above,2/5.57/above,3/5.49/above,4/5.38/above,5/4.83/above}
] table {\vctreezsrhundredngp};

\end{axis}
\end{tikzpicture}} \\
        \end{tabular}
    \end{adjustbox}
    \caption{Impact of NGP on IMP and VCTree for predicate classification when reducing VG's ground-truth. Blue lines show mR@100; orange show zR@100. Solid lines show mR and zsR w/o NGP; dotted show mR and zsR w/ NGP.}
    \label{fig:varying-sizes}
\end{figure}
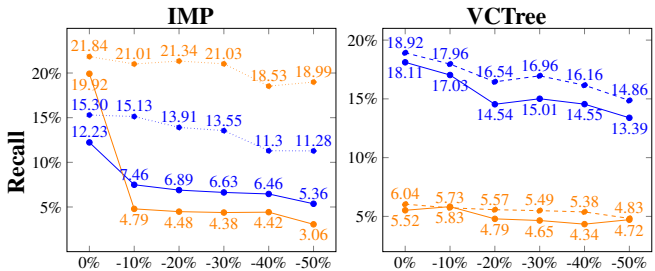

\conclude{NGP is particularly beneficial when reducing the amount of ground-truth facts.}
Figure~\ref{fig:varying-sizes} shows that the accuracy of SGG models can substantially decrease when reducing the ground-truth facts. In the case of VG, the most sensitive model is IMP: when reducing the training data by 50\%, zsR@100 drops by more than 6.5 times (19.92 \% vs. 3.06\%), while mR@100 drops by more than two times (12.23 \% vs. 5.36\%). In the case of OIv6, Table~\ref{tab:NGP-TDE-motifs-oiv6}, MOTIFS' mR@100 drops from 46.15\% to 42.17\% in predicate classification when reducing the ground-truth by 75\%; in scene graph classification, MOTIFS' mR@100 drops from 28.92\% to 23.28\%.
NGP can lead to drastic accuracy improvements for those cases.
Regarding IMP and VG, zsR@100 can increase from 3.06\% to 18.99\% when reducing the ground-truth by 50\%; zsR@100 can similarly increase from 5.36\% to 11.28\%. Similarly, when reducing by 75\% of the ground-truth of OIv6, mR@100 for predicate classification can increase from 42.17\% to 44.94\% in the case of MOTIFS; mR@100 can increase from 42.17\% to 44.94\% for scene graph classification, when NGP is applied.     

While NGP drops the mR of MOTIFS in scene graph classification when the whole ground-truth is used in OIv6, it is beneficial when reducing the ground-truth by 50\% and 75\%, Table~\ref{tab:NGP-TDE-motifs-oiv6}.
The high mR for MOTIFS even with significantly fewer ground-truth facts in Table~\ref{tab:NGP-TDE-motifs-oiv6} manifests that frequency-based techniques are effective for the OIv6 dataset.
Still, the integration with logic-based approaches (NGP) can further improve mR, Table~\ref{tab:NGP-TDE-motifs-oiv6}.
It is also worth noting that while TDE is particularly effective in VG, it decreases the mR of MOTIFS up to 11\% in OIv6. This is because OIv6 has a much higher annotation quality, and hence de-biasing is not crucial.
Finally, in contrast to NGP, TDE provides no supervision when reducing the ground-truth facts. 

\begin{figure}[htb]
\vspace{-0.95cm}
\centering
\includegraphics[width=0.9\columnwidth]{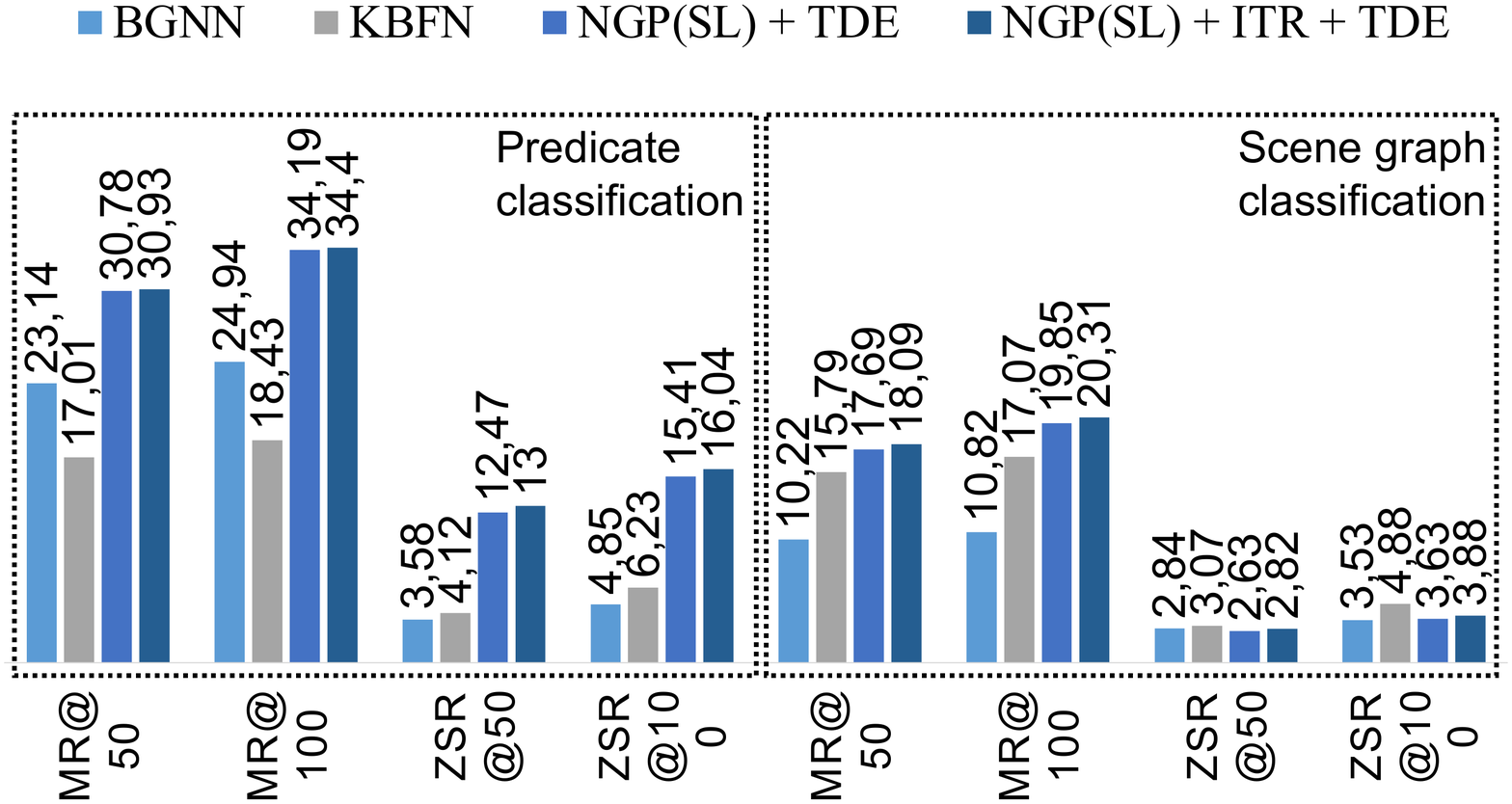} 
\vspace{-0.9cm}
\caption{Regularization vs. ad-hoc architectures and sophisticated models. Results on the VG dataset.}
\label{fig:kbfn}
\end{figure}

\conclude{Regularization can be more effective than sophisticated (neurosymbolic) SGG models.} 
The mR@k of KBFN is $17.01\%$ and $18.43\%$ for predicate classification, see Figure~\ref{fig:kbfn}. When jointly regularizing VCTree using NGP(SL) and TDE, the mR@k is $30.78\%$ and $34.19\%$. Similarly, for scene graph generation, the mR@k of KBFN is $15.79\%$ and $17.07\%$, and $17.69\%$ and $19.85\%$ for the regularized VCTree model.
Likewise, the regularized VCTree model reaches up to $90\%$ higher performance than BGNN.
These results show that regularizing a standard SGG model like VCTree, can be more effective than ad-hoc, neurosymbolic SGG architectures or more sophisticated models.

\section{Related work} \label{section:related}

Regularising neural models using symbolic knowledge has been extensively studied in information and natural language analysis \cite{DBLP:conf/aaai/WangP20,DBLP:conf/conll/Minervini018,rocktaschel-etal-2015-injecting}. Unlike the above line of research,
NGP focuses on scalable knowledge injection into SGG models under different semantics.

Differently from contrastive learning \cite{oord2018representation,chen2021exploring,technologies9010002} where models are trained in an unsupervised fashion by performing tasks that can be created from the input itself, NGP trains neural models using symbolic domain knowledge. 
The authors in \cite{EnergySGG} train a graph neural network to learn the joint conditional density of a scene graph and then use it as a loss function. 
To deal with the ambiguity in the SGG annotations, 
the work in \cite{DBLP:conf/cvpr/YangZ0WY21} generates different probabilistic representations of the predicates. In contrast to NGP, the above techniques do not support external knowledge.
Finally, the work in \cite{DBLP:conf/iccv/Zhong0YX021} generates localized scene graphs from image-text pairs; the technique does not rely on logic, but exclusively on neural models. 
Integrating logic-based regularization with the above research is an interesting future direction.  

Every technique that uses learned or fixed background knowledge as a prior, e.g., \cite{DBLP:conf/cvpr/GuZL0CL19,Zareian-ECCV-2020}, {is biased} towards that knowledge.
Differently from techniques like MOTIFS \cite{MOTIFS}, NGP is not biased by the frequency of the training facts: if the background knowledge is independent of the training facts or their frequencies, then NGP will not be biased toward the training facts or their frequencies. The above holds as both the logic-based losses and NGP's mechanism for choosing the maximally violated ICs are indifferent to any frequencies.
\section{Conclusions}
We introduced NGP, the first highly-scalable, symbolic, SGG regularization framework that leads to state-of-the-art accuracy. 
Future research includes supporting richer formulas and regularizing models under theories mined via knowledge extraction e.g., \cite{DBLP:conf/eccv/ZhuFF14}-- NGP supports such theories by weighting the ICs. Integrating NGP with neurosymbolic techniques that support indirect supervision like DeepProbLog \cite{deepproblog}, NeuroLog \cite{tsamoura2020neuralsymbolic} and ABL \cite{ABL} is another direction for future research.

\bibliography{references}

\begin{thebibliography}{43}
\providecommand{\natexlab}[1]{#1}

\bibitem[{Bach et~al.(2017)Bach, Broecheler, Huang, and Getoor}]{psl-long}
Bach, S.~H.; Broecheler, M.; Huang, B.; and Getoor, L. 2017.
\newblock Hinge-Loss Markov Random Fields and Probabilistic Soft Logic.
\newblock \emph{Journal of Machine Learning Research}, 18: 109:1--109:67.

\bibitem[{Chavira and Darwiche(2008)}]{wmc}
Chavira, M.; and Darwiche, A. 2008.
\newblock On probabilistic inference by weighted model counting.
\newblock \emph{Artificial Intelligence}, 172(6): 772 -- 799.

\bibitem[{Chen et~al.(2019)Chen, Yu, Chen, and Lin}]{KERN}
Chen, T.; Yu, W.; Chen, R.; and Lin, L. 2019.
\newblock Knowledge-embedded routing network for scene graph generation.
\newblock In \emph{Proceedings of the IEEE/CVF Conference on Computer Vision
  and Pattern Recognition}, 6163--6171.

\bibitem[{Chen and He(2021)}]{chen2021exploring}
Chen, X.; and He, K. 2021.
\newblock Exploring simple siamese representation learning.
\newblock In \emph{CVPR}, 15750--15758.

\bibitem[{Dai et~al.(2019)Dai, Xu, Yu, and Zhou}]{ABL}
Dai, W.-Z.; Xu, Q.; Yu, Y.; and Zhou, Z.-H. 2019.
\newblock {Bridging Machine Learning and Logical Reasoning by Abductive
  Learning}.
\newblock In \emph{NeurIPS}, 2815--2826.

\bibitem[{Dao et~al.(2021)Dao, Kamath, Syrgkanis, and Mackey}]{distillation2}
Dao, T.; Kamath, G.~M.; Syrgkanis, V.; and Mackey, L. 2021.
\newblock Knowledge Distillation as Semiparametric Inference.
\newblock In \emph{ICLR}.

\bibitem[{d'Avila Garcez, Broda, and Gabbay(2002)}]{DBLP:books/daglib/0007534}
d'Avila Garcez, A.~S.; Broda, K.; and Gabbay, D.~M. 2002.
\newblock \emph{Neural-symbolic learning systems: foundations and
  applications}.
\newblock Perspectives in neural computing. Springer.

\bibitem[{Donadello, Serafini, and d'Avila Garcez(2017)}]{LTN}
Donadello, I.; Serafini, L.; and d'Avila Garcez, A.~S. 2017.
\newblock Logic Tensor Networks for Semantic Image Interpretation.
\newblock In \emph{IJCAI}, 1596--1602.

\bibitem[{Fischer et~al.(2019)Fischer, Balunovic, Drachsler{-}Cohen, Gehr,
  Zhang, and Vechev}]{dl2}
Fischer, M.; Balunovic, M.; Drachsler{-}Cohen, D.; Gehr, T.; Zhang, C.; and
  Vechev, M.~T. 2019.
\newblock {DL2:} Training and Querying Neural Networks with Logic.
\newblock In \emph{ICML}, volume~97, 1931--1941.

\bibitem[{Gu et~al.(2019)Gu, Zhao, Lin, Li, Cai, and
  Ling}]{DBLP:conf/cvpr/GuZL0CL19}
Gu, J.; Zhao, H.; Lin, Z.; Li, S.; Cai, J.; and Ling, M. 2019.
\newblock Scene Graph Generation With External Knowledge and Image
  Reconstruction.
\newblock In \emph{CVPR}, 1969--1978.

\bibitem[{H{\'{a}}jek, Godo, and Esteva(2013)}]{fuzzy-probability}
H{\'{a}}jek, P.; Godo, L.; and Esteva, F. 2013.
\newblock Fuzzy Logic and Probability.
\newblock \emph{CoRR}, abs/1302.4953.

\bibitem[{Hinton, Vinyals, and Dean(2015)}]{distillation3}
Hinton, G.~E.; Vinyals, O.; and Dean, J. 2015.
\newblock Distilling the Knowledge in a Neural Network.
\newblock \emph{CoRR}, abs/1503.02531.

\bibitem[{Jaiswal et~al.(2021)Jaiswal, Babu, Zadeh, Banerjee, and
  Makedon}]{technologies9010002}
Jaiswal, A.; Babu, A.~R.; Zadeh, M.~Z.; Banerjee, D.; and Makedon, F. 2021.
\newblock A Survey on Contrastive Self-Supervised Learning.
\newblock \emph{Technologies}, 9(1).

\bibitem[{Kendall, Gal, and Cipolla(2018)}]{weighting}
Kendall, A.; Gal, Y.; and Cipolla, R. 2018.
\newblock Multi-Task Learning Using Uncertainty to Weigh Losses for Scene
  Geometry and Semantics.
\newblock In \emph{CVPR}.

\bibitem[{Krishna et~al.(2017)Krishna, Zhu, Groth, Johnson, Hata, Kravitz,
  Chen, Kalantidis, Li, Shamma, Bernstein, and Fei{-}Fei}]{VG}
Krishna, R.; Zhu, Y.; Groth, O.; Johnson, J.; Hata, K.; Kravitz, J.; Chen, S.;
  Kalantidis, Y.; Li, L.; Shamma, D.~A.; Bernstein, M.~S.; and Fei{-}Fei, L.
  2017.
\newblock Visual Genome: Connecting Language and Vision Using Crowdsourced
  Dense Image Annotations.
\newblock \emph{Int. J. Comput. Vis.}, 123(1): 32--73.

\bibitem[{Kuznetsova et~al.(2020)Kuznetsova, Rom, Alldrin, Uijlings, Krasin,
  Pont{-}Tuset, Kamali, Popov, Malloci, Kolesnikov, Duerig, and Ferrari}]{OIv4}
Kuznetsova, A.; Rom, H.; Alldrin, N.; Uijlings, J. R.~R.; Krasin, I.;
  Pont{-}Tuset, J.; Kamali, S.; Popov, S.; Malloci, M.; Kolesnikov, A.; Duerig,
  T.; and Ferrari, V. 2020.
\newblock The Open Images Dataset {V4}.
\newblock \emph{International Journal of Computer Vision}, 128(7): 1956--1981.

\bibitem[{Li et~al.(2021)Li, Zhang, Wan, and He}]{BGNN}
Li, R.; Zhang, S.; Wan, B.; and He, X. 2021.
\newblock Bipartite Graph Network With Adaptive Message Passing for Unbiased
  Scene Graph Generation.
\newblock In \emph{CVPR}, 11109--11119.

\bibitem[{Lu et~al.(2016)Lu, Krishna, Bernstein, and Fei-Fei}]{lu2016visual}
Lu, C.; Krishna, R.; Bernstein, M.; and Fei-Fei, L. 2016.
\newblock Visual Relationship Detection with Language Priors.
\newblock In \emph{ECCV}.

\bibitem[{Manhaeve et~al.(2018)Manhaeve, Dumancic, Kimmig, Demeester, and
  De~Raedt}]{deepproblog}
Manhaeve, R.; Dumancic, S.; Kimmig, A.; Demeester, T.; and De~Raedt, L. 2018.
\newblock DeepProbLog: Neural Probabilistic Logic Programming.
\newblock In \emph{NeurIPS}, 3749--3759.

\bibitem[{Minervini and Riedel(2018)}]{DBLP:conf/conll/Minervini018}
Minervini, P.; and Riedel, S. 2018.
\newblock Adversarially Regularising Neural {NLI} Models to Integrate Logical
  Background Knowledge.
\newblock In \emph{CoNLL}, 65--74.

\bibitem[{Oord, Li, and Vinyals(2018)}]{oord2018representation}
Oord, A. v.~d.; Li, Y.; and Vinyals, O. 2018.
\newblock Representation learning with contrastive predictive coding.
\newblock \emph{arXiv:1807.03748}.

\bibitem[{Ren et~al.(2015)Ren, He, Girshick, and Sun}]{NIPS2015_14bfa6bb}
Ren, S.; He, K.; Girshick, R.; and Sun, J. 2015.
\newblock Faster R-CNN: Towards Real-Time Object Detection with Region Proposal
  Networks.
\newblock In \emph{NeurIPS}.

\bibitem[{Rockt{\"a}schel, Singh, and
  Riedel(2015)}]{rocktaschel-etal-2015-injecting}
Rockt{\"a}schel, T.; Singh, S.; and Riedel, S. 2015.
\newblock Injecting Logical Background Knowledge into Embeddings for Relation
  Extraction.
\newblock In \emph{ACL}, 1119--1129.

\bibitem[{Sap et~al.(2019)Sap, Bras, Allaway, Bhagavatula, Lourie, Rashkin,
  Roof, Smith, and Choi}]{atomic}
Sap, M.; Bras, R.~L.; Allaway, E.; Bhagavatula, C.; Lourie, N.; Rashkin, H.;
  Roof, B.; Smith, N.~A.; and Choi, Y. 2019.
\newblock {ATOMIC:} An Atlas of Machine Commonsense for If-Then Reasoning.
\newblock In \emph{AAAI}, 3027--3035.

\bibitem[{Speer, Chin, and Havasi(2017)}]{conceptnet}
Speer, R.; Chin, J.; and Havasi, C. 2017.
\newblock ConceptNet 5.5: An Open Multilingual Graph of General Knowledge.
\newblock In \emph{AAAI}, 4444--4451.

\bibitem[{Suhail et~al.(2021)Suhail, Mittal, Siddiquie, Broaddus, Eledath,
  Medioni, and Sigal}]{EnergySGG}
Suhail, M.; Mittal, A.; Siddiquie, B.; Broaddus, C.; Eledath, J.; Medioni,
  G.~G.; and Sigal, L. 2021.
\newblock Energy-Based Learning for Scene Graph Generation.
\newblock In \emph{CVPR}, 13936--13945.

\bibitem[{Tang(2020)}]{tang2020sggcode}
Tang, K. 2020.
\newblock A Scene Graph Generation Codebase in PyTorch.
\newblock \url{https://github.com/KaihuaTang/Scene-Graph-Benchmark.pytorch}.

\bibitem[{Tang et~al.(2020)Tang, Niu, Huang, Shi, and Zhang}]{TDE}
Tang, K.; Niu, Y.; Huang, J.; Shi, J.; and Zhang, H. 2020.
\newblock Unbiased Scene Graph Generation From Biased Training.
\newblock In \emph{CVPR}, 3713--3722.

\bibitem[{Tang et~al.(2019)Tang, Zhang, Wu, Luo, and Liu}]{VCTREE}
Tang, K.; Zhang, H.; Wu, B.; Luo, W.; and Liu, W. 2019.
\newblock Learning to Compose Dynamic Tree Structures for Visual Contexts.
\newblock In \emph{CVPR}, 6619--6628.

\bibitem[{Tsamoura, Hospedales, and Michael(2021)}]{tsamoura2020neuralsymbolic}
Tsamoura, E.; Hospedales, T.; and Michael, L. 2021.
\newblock Neural-Symbolic Integration: A Compositional Perspective.
\newblock In \emph{AAAI}.

\bibitem[{{van Krieken}, Acar, and {van
  Harmelen}(2019)}]{van-Krieken-semi-supervised}
{van Krieken}, E.; Acar, E.; and {van Harmelen}, F. 2019.
\newblock Semi-Supervised Learning using Differentiable Reasoning.
\newblock \emph{IFCoLog Journal of Logic and its Applications}, 6(4): 633--653.

\bibitem[{van Krieken, Acar, and van Harmelen(2020)}]{KR2020-92}
van Krieken, E.; Acar, E.; and van Harmelen, F. 2020.
\newblock {Analyzing Differentiable Fuzzy Implications}.
\newblock In \emph{KR}, 893--903.

\bibitem[{Wang and Pan(2020)}]{DBLP:conf/aaai/WangP20}
Wang, W.; and Pan, S.~J. 2020.
\newblock Integrating Deep Learning with Logic Fusion for Information
  Extraction.
\newblock In \emph{AAAI}, 9225--9232.

\bibitem[{Xie et~al.(2019)Xie, Xu, Meel, Kankanhalli, and Soh}]{LENSR}
Xie, Y.; Xu, Z.; Meel, K.~S.; Kankanhalli, M.~S.; and Soh, H. 2019.
\newblock Embedding Symbolic Knowledge into Deep Networks.
\newblock In \emph{NeurIPS}, 4235--4245.

\bibitem[{{Xu} et~al.(2017){Xu}, {Zhu}, {Choy}, and {Fei-Fei}}]{IMP}
{Xu}, D.; {Zhu}, Y.; {Choy}, C.~B.; and {Fei-Fei}, L. 2017.
\newblock Scene Graph Generation by Iterative Message Passing.
\newblock In \emph{{CVPR}}, 3097--3106.

\bibitem[{Xu et~al.(2018)Xu, Zhang, Friedman, Liang, and Van~den
  Broeck}]{semantic-loss}
Xu, J.; Zhang, Z.; Friedman, T.; Liang, Y.; and Van~den Broeck, G. 2018.
\newblock A Semantic Loss Function for Deep Learning with Symbolic Knowledge.
\newblock In \emph{ICML}, 5502--5511.

\bibitem[{Yang et~al.(2021)Yang, Zhang, Zhang, Wu, and
  Yang}]{DBLP:conf/cvpr/YangZ0WY21}
Yang, G.; Zhang, J.; Zhang, Y.; Wu, B.; and Yang, Y. 2021.
\newblock Probabilistic Modeling of Semantic Ambiguity for Scene Graph
  Generation.
\newblock In \emph{CVPR}, 12527--12536.

\bibitem[{Zareian, Karaman, and Chang(2020)}]{Zareian_2020_ECCV}
Zareian, A.; Karaman, S.; and Chang, S.-F. 2020.
\newblock Bridging Knowledge Graphs to Generate Scene Graphs.
\newblock In \emph{ECCV}.

\bibitem[{Zareian et~al.(2020)Zareian, Wang, You, and
  Chang}]{Zareian-ECCV-2020}
Zareian, A.; Wang, Z.; You, H.; and Chang, S.-F. 2020.
\newblock Learning Visual Commonsense for Robust Scene Graph Generation.
\newblock In \emph{ECCV}, 642--657.

\bibitem[{Zellers et~al.(2018)Zellers, Yatskar, Thomson, and Choi}]{MOTIFS}
Zellers, R.; Yatskar, M.; Thomson, S.; and Choi, Y. 2018.
\newblock Neural Motifs: Scene Graph Parsing With Global Context.
\newblock In \emph{CVPR}, 5831--5840.

\bibitem[{Zhang et~al.(2017)Zhang, Kyaw, Chang, and Chua}]{VTranceE}
Zhang, H.; Kyaw, Z.; Chang, S.; and Chua, T. 2017.
\newblock Visual Translation Embedding Network for Visual Relation Detection.
\newblock In \emph{CVPR}, 3107--3115.

\bibitem[{Zhong et~al.(2021)Zhong, Shi, Yang, Xu, and
  Li}]{DBLP:conf/iccv/Zhong0YX021}
Zhong, Y.; Shi, J.; Yang, J.; Xu, C.; and Li, Y. 2021.
\newblock Learning to Generate Scene Graph from Natural Language Supervision.
\newblock In \emph{ICCV}, 1803--1814.

\bibitem[{Zhu, Fathi, and Fei{-}Fei(2014)}]{DBLP:conf/eccv/ZhuFF14}
Zhu, Y.; Fathi, A.; and Fei{-}Fei, L. 2014.
\newblock Reasoning about Object Affordances in a Knowledge Base
  Representation.
\newblock In \emph{ECCV}, volume 8690, 408--424.

\end{thebibliography}

\clearpage
\appendix
\section{Appendix}
\section{Adapting Logic Tensor Networks for our analysis}\label{appendix:ltns}

LTNs is a neurosymbolic framework \cite{LTN}. 
The framework was applied to the task of predicting whether the objects enclosed in specific bounding boxes  
adhere to the \texttt{partOf} relation. 
In this section, we present how we extended LTNs for predicate and scene graph classification in our experimental setup.

\paragraph{Background} 
LTNs consists of two parts: a neural model $n$ for object classification and, on top of $n$,  
a logical theory $T$ that reasons in a symbolic fashion over the predictions of $n$. 
In the setting presented in \cite{LTN}, theory $T$ outputs facts of the form 
$\texttt{o}(b)$ and $\texttt{partOf}(b_1,b_2)$ denoting that the object within bounding box $b$ is of type $\texttt{o}$ and 
that the object within $b_1$ is a part of the object within $b_2$, respectively. 
The semantics of $T$ is defined through a class of interpretation functions $\mathcal{G}$  
mapping each fact to the ${[0,1]}$ interval. 
In particular, the confidence of a fact $\texttt{o}(b)$ is given by $\mathcal{G}(\texttt{o})(b)$, while the confidence of  
a fact $\texttt{partOf}(b_1,b_2)$ is given by $\mathcal{G}(\texttt{partOf})(b_1,b_2)$, where 
$\mathcal{G}(\texttt{o})$ and $\mathcal{G}(\texttt{partOf})$ are functions from 
bounding boxes\footnote{LTNs represent bounding boxes using their upper left and lower right coordinates in the image.} and pairs of bounding boxes into $[0,1]$.
Intuitively, the functions in $\mathcal{G}$ reflect the degree to which the \emph{entire} framework 
has certain beliefs on the types of the objects or their interrelationships. 
To accommodate logical theories, $\mathcal{G}$ additionally includes functions for interpreting formulas in first order logic using the semantics of Lukasiewicz's fuzzy logic: 
\begin{align}
    \mathcal{G}(\neg \varphi) &\defeq 1 - \mathcal{G}(\varphi) \\
    \mathcal{G}(\varphi_1 \wedge \varphi_2) &\defeq \max\{0, \mathcal{G}(\varphi_1) + \mathcal{G}(\varphi_2) -1\} \\
    \mathcal{G}(\varphi_1 \vee \varphi_2) &\defeq \min\{1, \mathcal{G}(\varphi_1) + \mathcal{G}(\varphi_2)\} 
\end{align}

Above, $\varphi$, $\varphi_1$ 
and $\varphi_2$ are formulas over variables and the Boolean connectives $\wedge$, $\vee$ and $\neg$.
The aim of LTNs is to learn the weights of the neural model so that $\mathcal{G}$ correctly predicts (i) the object within a bounding box, as well as (ii) whether two objects described in terms of their surrounding bounding boxes abide by the \texttt{partOf} relation. Regularization of the neural model is achieved via $\mathcal{G}$, in the sense that the model's weights are changed so that the outputs of $\mathcal{G}$ satisfy the background knowledge as well as agree with the annotations in the ground truth. Below, we discuss how we used LTNs for predicate and scene graph classification. 
Before that, notice that the neural SGG models provide two interfaces. Given a bounding box $b$, the first interface outputs the confidence to which the object enclosed in $b$ belongs to a class $\texttt{o}$ in $\texttt{O}$. Given two bounding boxes $b$ and $b^{\prime}$, the second interface outputs the confidence to which the objects enclosed in $b$ and $b^{\prime}$ relate according to a predicate $\texttt{p}$ in \texttt{P}.
Furthermore, to establish a fair comparison 
we used the neural model from \cite{NIPS2015_14bfa6bb} for object detection. 

We are now ready to describe our extension. 
The first step is to extend $\mathcal{G}$ with functions for object and predicate classification. 
These functions classify an object (resp. pair of objects) to an object class $\texttt{o}$ (resp. predicate class $\texttt{p}$), if $\texttt{o}$ (resp. $\texttt{p}$) is assigned the maximum confidence by the neural model. 
In particular, following \cite{LTN}, we added to $\mathcal{G}$ a function $\mathcal{G}(\texttt{o}_i)$, for each $\texttt{o}_i \in \texttt{O}$, which outputs 1 if the object enclosed in the input bounding box is of type $\texttt{o}_i$, and 0, otherwise. 
Furthermore, for each ${\texttt{p}_i \in \texttt{P}}$, we added a function $\mathcal{G}(\texttt{p}_i)$, which, given a pair of bounding boxes $b$ and $b'$, takes value 1 if the most likely predicate describing the relationship between $b$ and $b'$ is $\texttt{p}_i$ according to the neural model, and 0, otherwise. 

The second step is to add the background theory, while 
the final step is to create training data\footnote{The VG benchmark annotates each pair of bounding boxes $(b_i,b_j)$ in the ground truth with the types $\texttt{o}_i$ and $\texttt{o}_j$ of the enclosed objects, and the predicate \texttt{p} describing their relationship.}. We followed the procedure described in \cite{LTN}. For each bounding box $b_i$ annotated with the object class $\texttt{o}_i$, we added the fact $\texttt{o}_i(b_i)$, as well as the facts $\neg \texttt{o}_j(b_i)$, for each ${j \neq i}$. Furthermore, for each pair of bounding boxes $(b_i,b_j)$ annotated with predicate \texttt{p}, we added the fact $\texttt{p}(b_i,b_j)$, as well as the facts $\neg \texttt{p}^{\prime}(b_i,b_j)$, for each ${\texttt{p}^{\prime} \neq \texttt{p}}$.
All facts in the training set have confidence 1. Provided with all this information, LTNs train the neural SGG model.  
\section{Implementation Details \& Additional Results}\label{appendix:extra-results}

\paragraph{Implementation details} 
We used the state of the art open source library from \cite{tang2020sggcode} 
(released under the MIT license) for implementing all neural SGG models considered in our evaluation.
The library fixes bugs affecting previous implementations that were leading to very high results. This is why the reported results may differ from the ones in previously published work.

To ensure a fair comparison and following the widely adopted protocol in literature, we used the same pre-trained Faster R-CNN \cite{NIPS2015_14bfa6bb} backbone for object detection\footnote{For predicate and scene graph classification, Faster R-CNN acts only as a feature extractor.} for all models following the procedure from \cite{TDE} for training it. 
We train the SGG models of IMP, MOTIFS and VCTree using SGD, with a batch size of 12, a learning rate of $1 \times 10^{-2}$, and a weight decay of $1 \times 10^{-4}$, keeping the hyperparameters recommended by the authors. 

For the experiments with OIv6, we used for all models the same pre-trained Faster R-CNN with the experiments with VG. The hyperparameters presented above are also used for the OIv6 dataset, following the evaluation procedure by \cite{BGNN}.
To evaluate BGNN, we used the implementation\footnote{Available at \url{https://github.com/Scarecrow0/BGNN-SGG} under the MIT license.} provided by the authors \cite{BGNN}. It should be stressed that the results reported for BGNN in \cite{BGNN} 
are not reproducible with the codebase provided by the authors. This is an ongoing issue that has been also reported by the other users of the codebase.   

To compute SL, we used the PySDD library version 0.2.10 (licensed under the Apache License, version 2.0) that compiles formulas into arithmetic circuits.

\subsection{Additional results}

\begin{table*}[t]
    \centering
    \caption{Impact of the number of ICs on NGP's accuracy. Results on the VG dataset.}
    \label{tab:different_k}
    \resizebox{\textwidth}{!}{%
    \begin{tabular}{lllrccccccc|ccccccc}
        \toprule
        Model & Theory & Regularization & \# ICs & \multicolumn{7}{c}{Predicate Classification} & \multicolumn{7}{c}{Scene Graph Classification} \\
        & & & & \multicolumn{3}{c}{mR@} & \multicolumn{3}{c}{zsR@} & Time (s) & \multicolumn{3}{c}{mR@} & \multicolumn{3}{c}{zsR@} & Time (s)\\
        & & & & 20 & 50 & 100 & 20 & 50 & 100 & & 20 & 50 & 100 & 20 & 50 & 100 &\\
        \midrule
         VCTree & -              & TDE & 0 & 19.40 & 25.94 & 29.48 & 8.14 & 12.38 & 14.07 & 330.85 & 10.51 & 14.53 & 16.73 & 1.48 & 2.54 & 3.99 & 370.22\\
         VCTree & $\theorya$     & NGP(SL)+TDE & 2 & 23.91 & 30.78 & 34.19 & \textbf{8.15} & \textbf{12.47} & \textbf{15.41} & 367.14 & \textbf{13.60} & \textbf{17.69} & \textbf{19.85} & 1.57 & 2.63 & 3.63 & 437.44\\
         VCTree & $\theorya$     & NGP(SL)+TDE & 3 & 23.99 & 31.31 & 35.10 & 6.72 & 10.61 & 13.36 & 389.29 & 13.18 & 17.23 & 19.42 & 1.67 & 3.00 & 3.95 & 617.89\\
         VCTree & $\theorya$     & NGP(SL)+TDE & 5 & 23.90 & 31.22 & 35.17 & 6.81 & 10.69 & 13.24 & 487.04 & {13.50} & {17.31} & {19.70} & \textbf{2.74} & \textbf{4.09} & \textbf{5.12} & 725.62\\
         VCTree & $\theorya$     & NGP(SL)+TDE & 7 & \textbf{24.75} & \textbf{32.14} & \textbf{35.82} & 7.17 & 10.81 & 13.68 & 540.01 & 12.57 & 17.22 & 19.33 & 1.64 & 2.93 & 3.59 & 834.85\\
         VCTree & $\theorya$     & NGP(SL)+TDE & 10 &  24.32 & 31.59 & 35.11 & 6.10 & 10.52 & 13.26 & 606.95 & 12.29 & 17.15 & 19.26 & 1.68 & 2.94 & 3.83 & 983.34\\
        \bottomrule
    \end{tabular}
    }
    \vspace{2pt}
    \centering
    \caption{Impact of randomly chosen ICs. Results on the VG dataset.}
    \label{tab:random_constraints}
    \resizebox{\textwidth}{!}{%
    \begin{tabular}{lllrcccccc|cccccc}
        \toprule
        Model &  Theory & Regularization & \# ICs & \multicolumn{6}{c}{Predicate Classification} & \multicolumn{6}{c}{Scene Graph Classification} \\
        & & & & \multicolumn{3}{c}{mR@} & \multicolumn{3}{c}{zsR@} & \multicolumn{3}{c}{mR@} & \multicolumn{3}{c}{zsR@} \\
         & & & & 20 & 50 & 100 & 20 & 50 & 100 & 20 & 50 & 100 & 20 & 50 & 100 \\
        \midrule
         VCTree   & $\theorya$   & TDE & 0 & 19.40 & 25.94 & 29.48 & \textbf{8.14} & \textbf{12.38} & \textbf{14.07} & 10.51 & 14.53 & 16.73 & 1.48 & 2.54 & 3.99\\
         VCTree   & $\theorya$    & NGP(SL)+TDE & 2 & 23.86 & 31.14 & 34.66 & 6.53 & 10.61 & 13.13 & 11.97 & 16.23 & 18.53 & 3.54 & \textbf{5.46} & 6.81\\
         VCTree   & $\theorya$    & NGP(SL)+TDE & 3 & 23.99 & 31.33 & 34.91 & 6.57 & 10.63 & 13.70 & 12.60 & 16.61 & 19.08 & 3.53 & 5.27 & 6.87\\
         VCTree   & $\theorya$    & NGP(SL)+TDE & 5 & \textbf{24.31} & \textbf{31.46} & \textbf{35.25} & 6.57 & 10.29 & 13.03 & \textbf{12.76} & \textbf{16.94} & \textbf{19.30} & \textbf{3.75} & 5.37 & \textbf{7.00}\\
         VCTree   & $\theorya$    & NGP(SL)+TDE & 7 & 23.63 & 30.85 & 34.58 & 6.57 & 10.38 & 13.14 & 10.92 & 15.30 & 17.71 & 3.04 & 4.93 & 6.47\\
         VCTree   & $\theorya$    & NGP(SL)+TDE & 10 & 10.93 & 14.16 & 15.45 & 5.16 & 9.90 & 13.49 & 10.90 & 15.27 & 17.67 & 3.02 & 4.91 & 6.49\\
        \bottomrule
    \end{tabular}
    }
    \vspace{2pt}
    \caption{Impact of GLAT and NGP on VCTree with TDE. Results on the VG dataset.}
    \label{tab:NGP-tde-glat}
    \resizebox{\textwidth}{!}{%
    \begin{tabular}{lllcccccc|cccccc}
        \toprule
        Model & Theory & Regularization & \multicolumn{6}{c}{Predicate Classification} & \multicolumn{6}{c}{Scene Graph Classification} \\ 
        & & & \multicolumn{3}{c}{mR@} & \multicolumn{3}{c}{zsR@} & \multicolumn{3}{c}{mR@} & \multicolumn{3}{c}{zsR@} \\
         & & & 20 & 50 & 100 & 20 & 50 & 100 & 20 & 50 & 100 & 20 & 50 & 100\\
        \midrule
        VCTree       & - & TDE & {19.40} & {25.94} & {29.48} & \textbf{{8.14}} & \textbf{{12.38}} & \textbf{{14.07}} & 10.51 &  14.53 & 16.73 & 1.48 & 2.54 & \textbf{{3.99}} \\
        VCTree & $\theoryb$  & NGP(SL)+TDE & \textbf{24.07} & \textbf{31.06} & \textbf{34.53} & 6.30 & 10.46 & 12.90 & \textbf{11.19} & \textbf{15.10} & \textbf{17.66} &\textbf{1.66} & \textbf{2.64} & {3.51} \\
        \midrule
        VCTree  & - & TDE & \textbf{19.40} & \textbf{25.94} & \textbf{29.48} & \textbf{8.14} & \textbf{12.38} & \textbf{14.07} & \textbf{10.51} &  \textbf{14.53} & \textbf{16.73} & \textbf{1.48} & \textbf{2.54} & \textbf{3.99} \\
        VCTree & -  & GLAT+TDE & 13.07 & 19.05 & 23.14 & 4.99 & 8.09 & 11.01 & 0.90 & 2.06 & 3.60 & 1.30 & 2.22 & 3.18 \\
        \bottomrule
    \end{tabular}
    }
    \vspace{2pt}
    \caption{Impact of different regularization techniques on BGNN's accuracy. Results on the VG dataset.}
    \label{tab:BGNN-LENSR-GLAT}
    \resizebox{\textwidth}{!}
    {%
    \begin{tabular}{lllcccccc|cccccc}
        \toprule
        Model & Theory & Regularization & \multicolumn{6}{c}{Predicate Classification} & \multicolumn{6}{c}{Scene Graph Classification} \\
        & & & \multicolumn{3}{c}{mR@} & \multicolumn{3}{c}{zsR@} & \multicolumn{3}{c}{mR@} & \multicolumn{3}{c}{zsR@} \\
        & & & 20 & 50 & 100 & 20 & 50 & 100 & 20 & 50 & 100 & 20 & 50 & 100 \\
        \midrule
        BGNN & - & - & \textbf{19.07} & \textbf{23.14} & \textbf{24.94} & \textbf{1.86} & \textbf{3.58} & \textbf{4.85} & \textbf{8.7} & \textbf{10.22} & \textbf{10.82} & \textbf{1.71} & \textbf{2.84} & \textbf{3.53} \\
	    BGNN & - & GLAT & \textbf{19.07} & \textbf{23.14} & \textbf{24.94} & \textbf{1.86} & \textbf{3.58} & \textbf{4.85} & 5.53 & 7.19 & 8.06 & 0.66 & 1.14 & 1.69 \\
	    \midrule
        BGNN & - & - & \textbf{19.07} & \textbf{23.14} & \textbf{24.94} & 1.86 & \textbf{3.58} & 4.85 & 8.7 & 10.22 & 10.82 & \textbf{1.71} & \textbf{2.84} & \textbf{3.53} \\
        BGNN & $\theorya$ & LENSR & 18.55 & 22.74 & 24.50 & \textbf{1.98} & 3.44 & \textbf{4.86} & \textbf{10} & \textbf{12.44} & \textbf{13.39} & 1.41 & 2.04 & 2.58 \\
        \midrule
	    BGNN & - & - & \textbf{19.07} & \textbf{23.14} & \textbf{24.94} & {1.86} & {3.58} & {4.85} & {8.7} & {10.22} & {10.82} & \textbf{1.71} & \textbf{2.84} & \textbf{3.53} \\
        BGNN & $\theorya$ & NGP(SL) & 17.41 & 21.38 & 22.91 & \textbf{2.43} & \textbf{4.52} & \textbf{6.53} & \textbf{9.39} & \textbf{11.23} & \textbf{11.98} & 1.66 & 2.78 & \textbf{3.53} \\
        \bottomrule
    \end{tabular}
    }
\end{table*}

\paragraph{NGP is time-efficient} 
Table~\ref{tab:different_k} reports results on VCTree with TDE when increasing the number of the ICs used by NGP. The benchmark is VG. We used this model combination as it leads to the highest recall in our previous analysis.  
The Time column reports the time in seconds for processing 200 batches of size 12 using two NVidia GeForce GTX 1080 Ti GPUs.
Table~\ref{tab:different_k} shows that the overhead to compute 
\eqref{eq:constraints} at each training step is small in practice: the runtime tends to increase linearly with the number $\rho$ of ICs. 

Table~\ref{tab:different_k} shows that both for  predicate classification and scene graph classification, the recall may drop by increasing $\rho$. 
We conjecture that this is due to the extreme bias 
and inaccuracies in the ground-truth in VG, a phenomenon already known to the community \cite{TDE}. In an unbiased dataset, the recall is expected to increase when the knowledge becomes richer (e.g., by increasing the number of ICs in our case) as the supervision signal becomes stronger. However, in the case of extreme bias, a more accurate model may not have the best recall. For instance, in quite a few cases, the ground-truth facts use the predicate \texttt{on}, despite that \texttt{laying on} better describes the relationship between the subject and the object. If a model is more accurate towards detecting the \texttt{laying on} relation, then some detections of \texttt{laying on} will be mistakenly considered as wrong ones dropping the recall measures.

\paragraph{Randomly chosen ICs can drop recall}
Table~\ref{tab:random_constraints} repeats the experiment from Table~\ref{tab:different_k},
by randomly choosing the ICs this time, though. The random selection technique 
leads to generally lower recall than the greedy technique proposed in Algorithm~\ref{algorithm:greedy}, while for zero-shot recall, the recall is consistently lower than that reported in Table~\ref{tab:different_k}. This highlights the effectiveness of our proposed strategy.

\paragraph{NGP is effectively integrated with TDE} Table~\ref{tab:NGP-tde-glat} reports results for NGP(SL) and GLAT on VCTree in combination with TDE. The benchmark is VG. 
In contrast to Table~\ref{tab:NGP-tde-conceptnet}, NGP is applied using $\theoryb$ to establish a fair comparison against GLAT by using knowledge 
exclusively coming from the training data. We can see that, in contrast to NGP, GLAT can substantially drop the recall of the model when employed in conjunction with TDE. For instance, mR@k drops from 19.40$\%$, 25.94$\%$ and 29.48$\%$ to 13.07$\%$, 19.05$\%$ and 23.14$\%$. In contrast to GLAT, NGP improves the recall of the model in most cases. Regarding predicate classification, mR@k increases to 24.07$\%$, 31.06$\%$ and 34.53$\%$, while regarding scene graph classification, mR@k increases to 11.19$\%$, 15.10$\%$ and 17.66$\%$.
We plan to investigate the cases where zsR@k drops for predicate classification.

\paragraph{BGNN is sensitive to regularization}
As discussed in the main body of the paper, the sampling-based approach of 
BGNN makes its integration with regularization based techniques difficult. 
Table~\ref{tab:BGNN-LENSR-GLAT} presents results on the integration of BGNN with GLAT, LENSR and NGP. 
We can see that GLAT has no impact on BGNN in predicate classification, while it drops its accuracy in scene graph classification. 
LENSR, in turn, drops both the mR and zsR for predicate classification in most cases improving only mR for scene graph classification. 
Overall, NGP is the most effective regularization technique as shown in Table~\ref{tab:BGNN-LENSR-GLAT}: it leads to substantial improvements in zsR and mR in predicate and scene graph classification, respectively, dropping the recall in the remaining cases, though.

\begin{figure}[h]
\centering
 \includegraphics[width=\columnwidth]{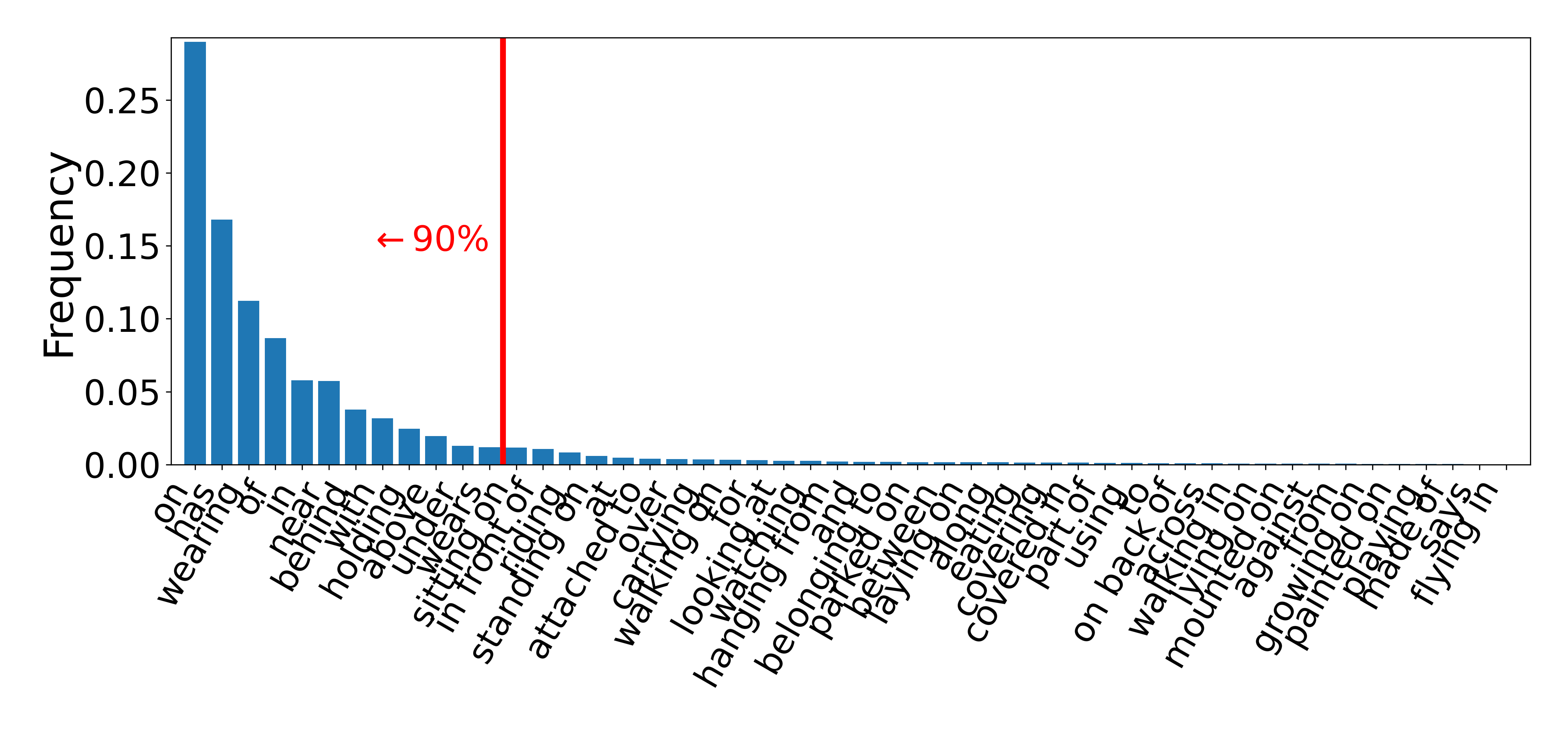}
\caption{Frequency of the ground-truth facts in VG grouped by their predicates.}
\label{Figure: Empirical Results 1}
\end{figure}

\paragraph{NGP improves recall for less frequent predicates}
Figure~\ref{Figure: Empirical Results 1} shows the frequency of ground-truth facts in VG grouped by their predicates.
The long tail effect becomes immediately apparent as the majority of predicates is used in less than 10\% of the ground-truth. For instance, the number of ground-truth facts having the \texttt{on} and \texttt{has} predicates is substantially higher than the number of facts having the \texttt{flying on} predicate. 

\begin{figure*}[t]
\centering
\begin{tabular}{c}
 \includegraphics[width=0.65\textwidth]{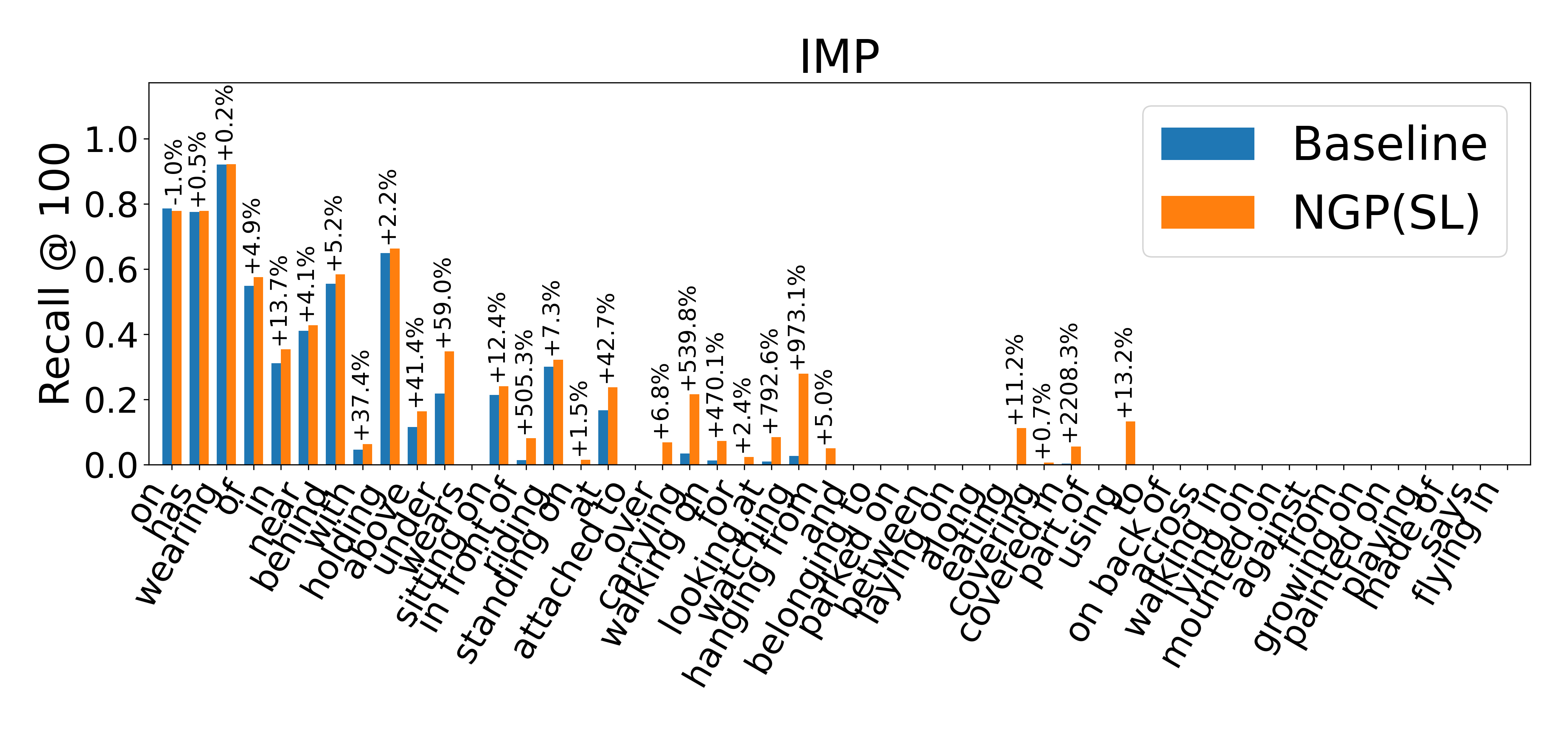} \\
 \includegraphics[width=0.65\textwidth]{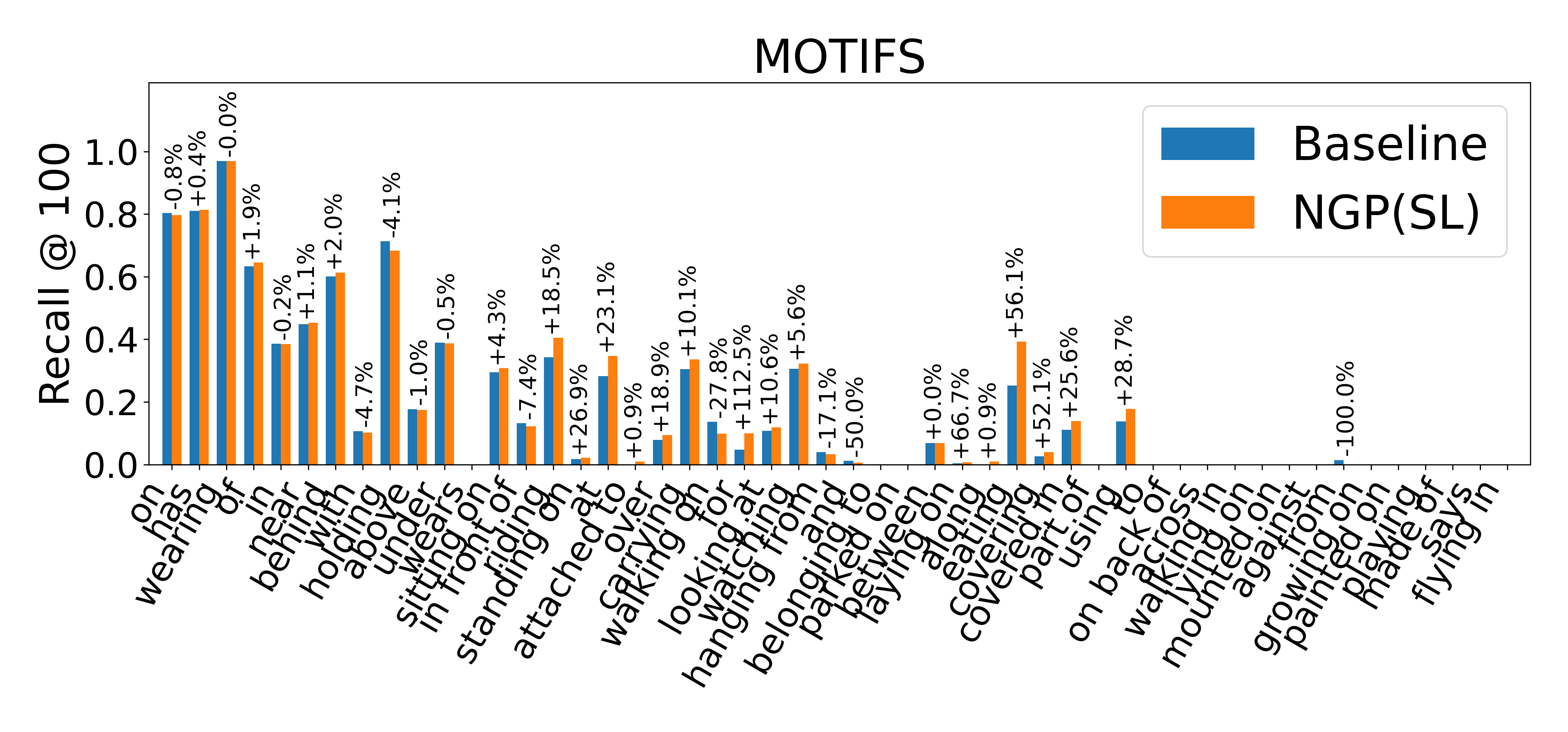} \\
 \includegraphics[width=0.65\textwidth]{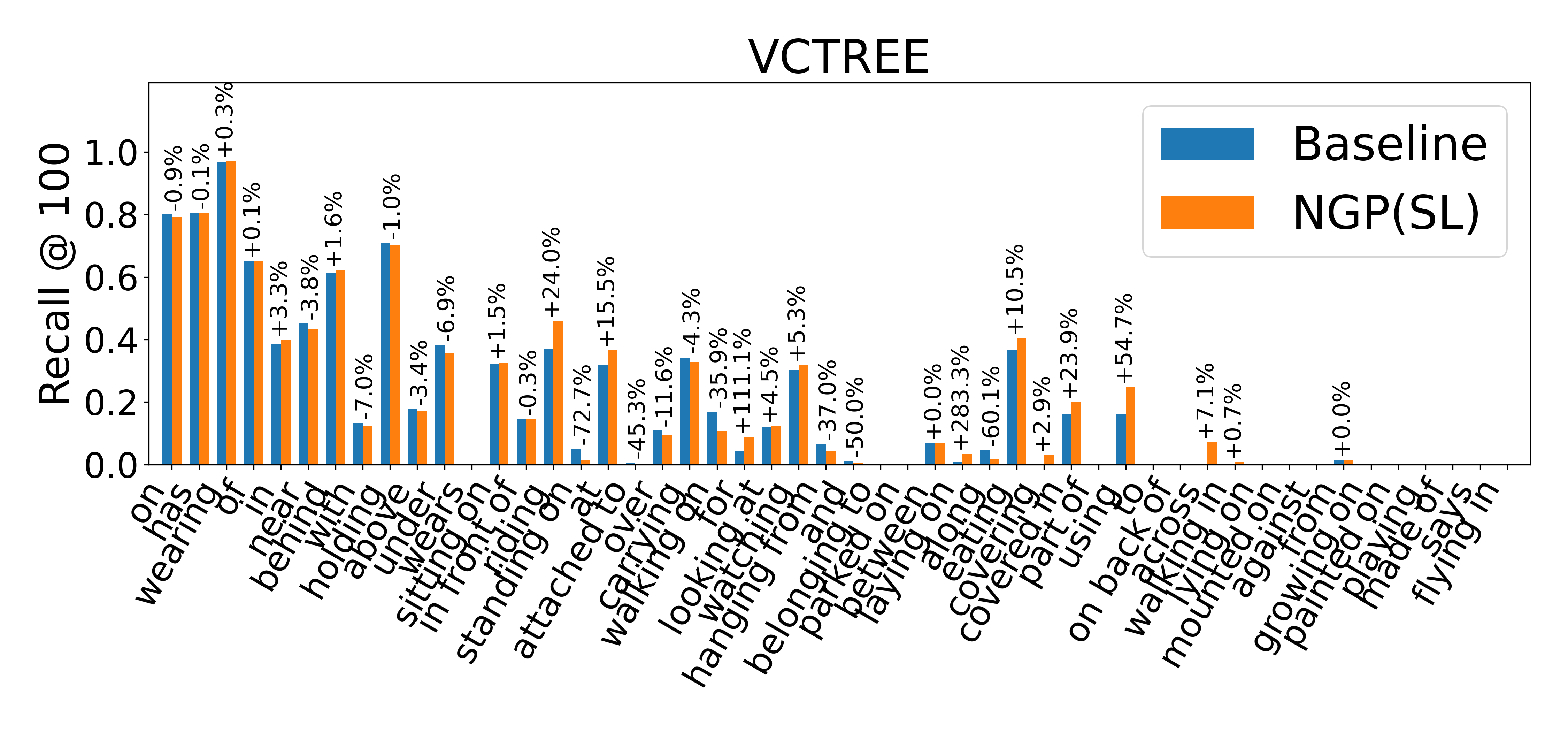} \\
 \includegraphics[width=0.65\textwidth]{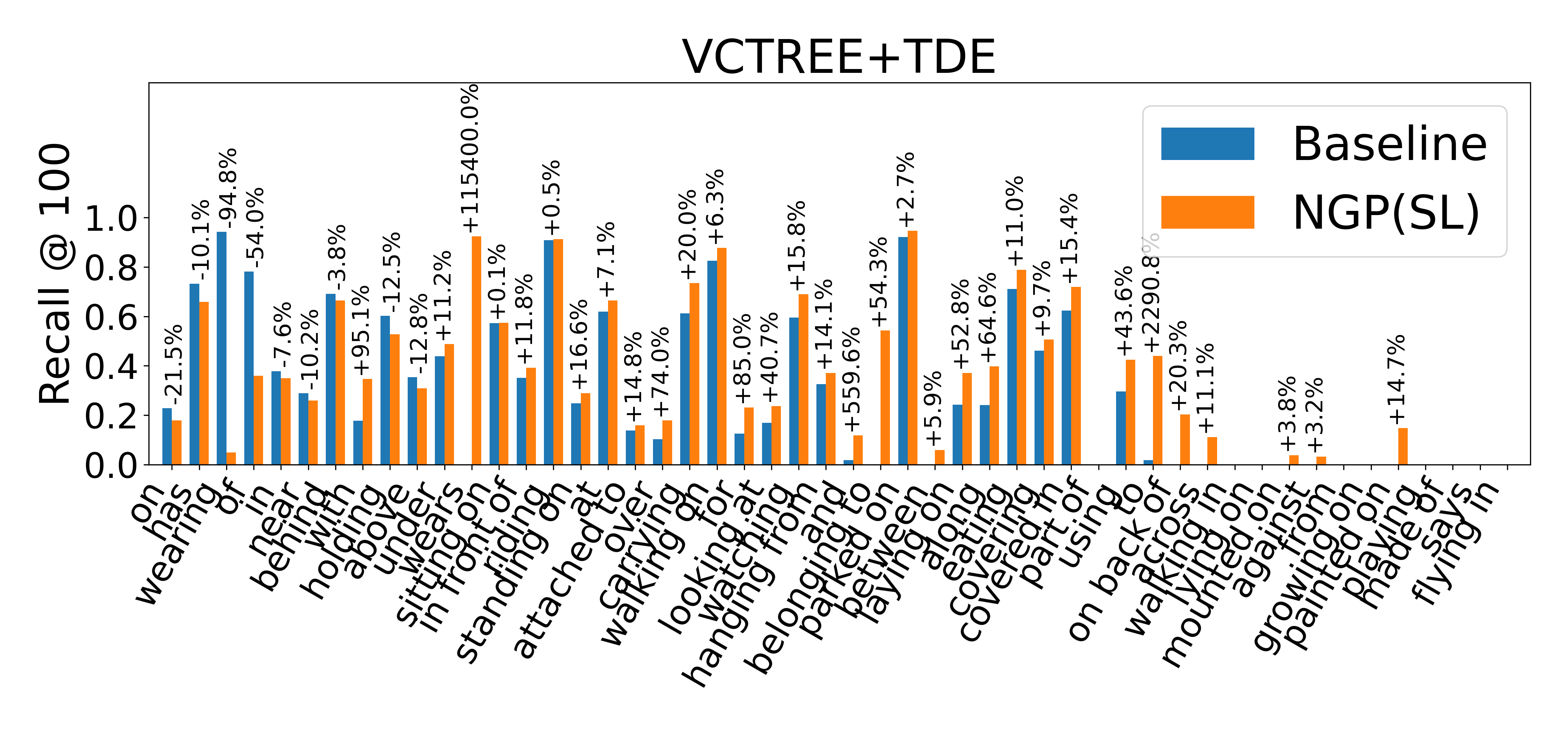} \\
\end{tabular}
\caption{Recall for predicate classification on a per-predicate basis. Results on the VG dataset.}
\label{Figure: Empirical Results 3}
\end{figure*}

\begin{figure*}[t]
\centering
\begin{tabular}{c}
 \includegraphics[width=0.65\textwidth]{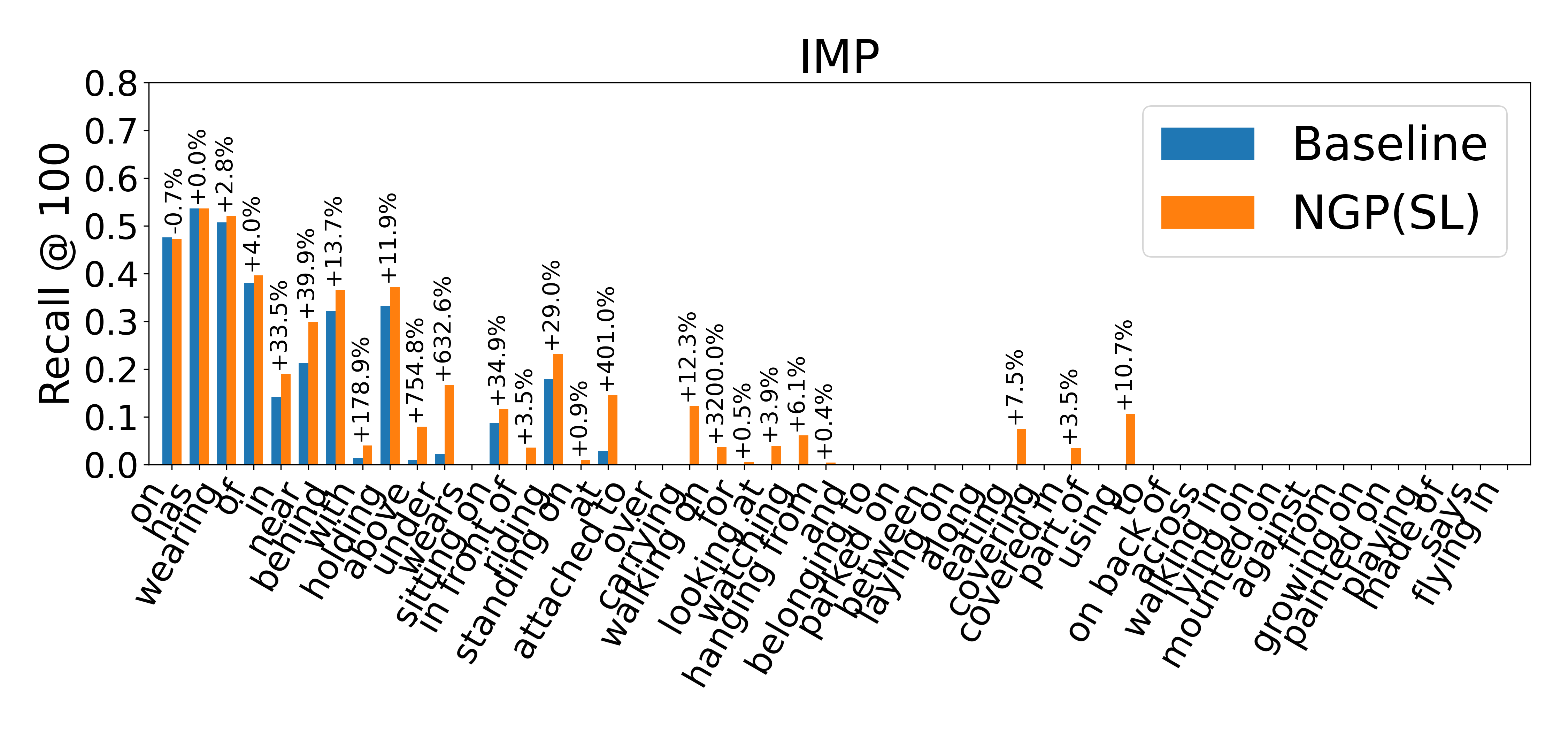} \\
 \includegraphics[width=0.65\textwidth]{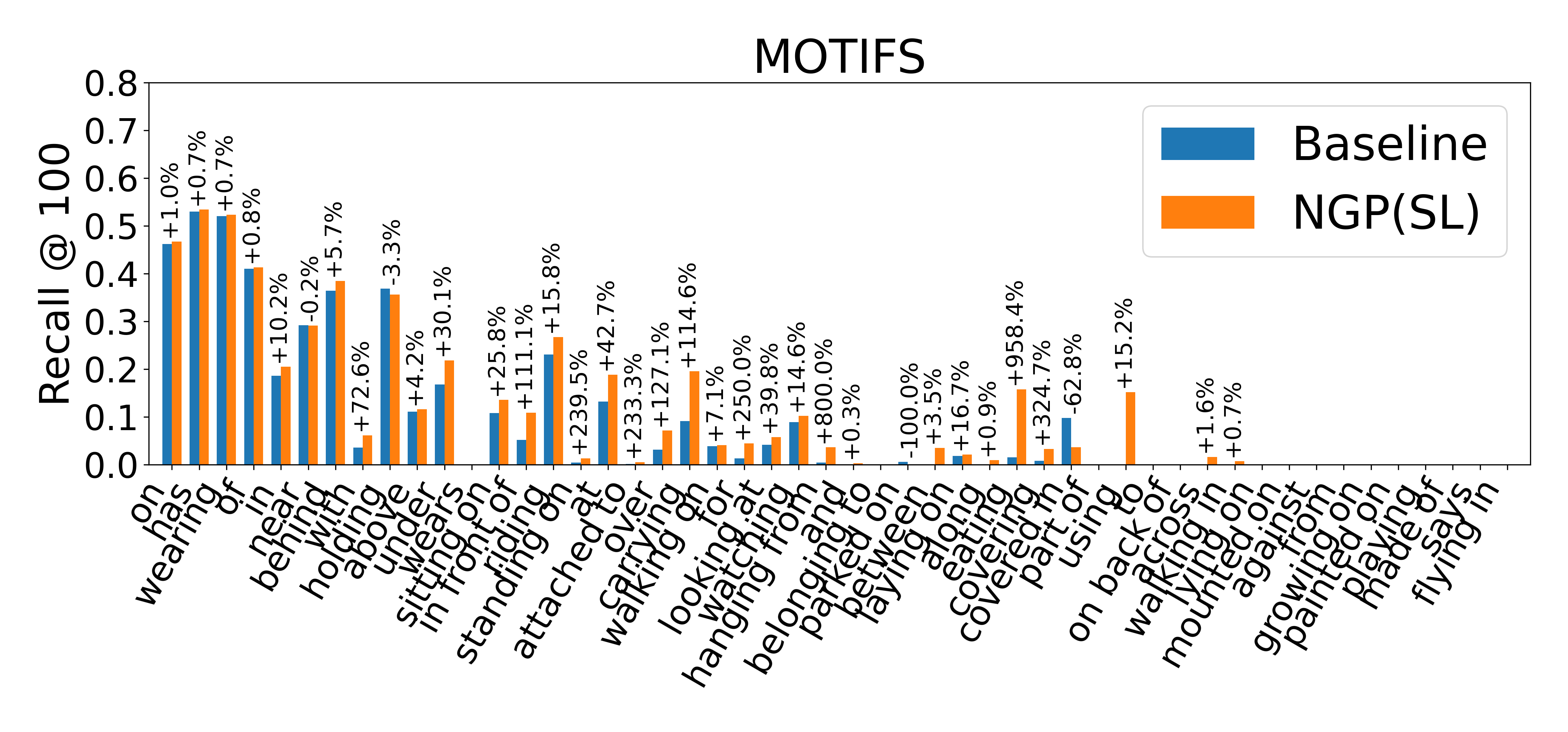} \\
 \includegraphics[width=0.65\textwidth]{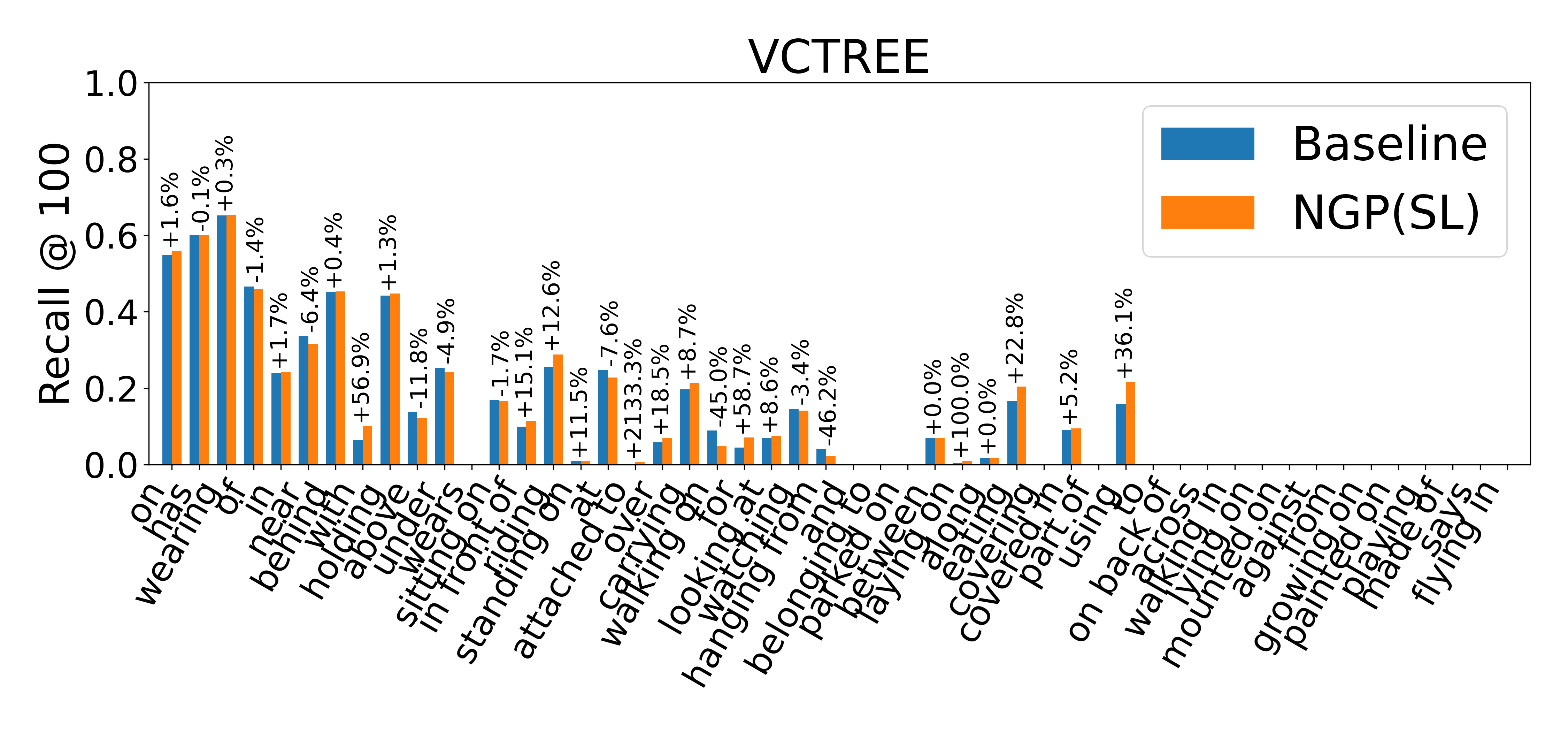} \\
 \includegraphics[width=0.65\textwidth]{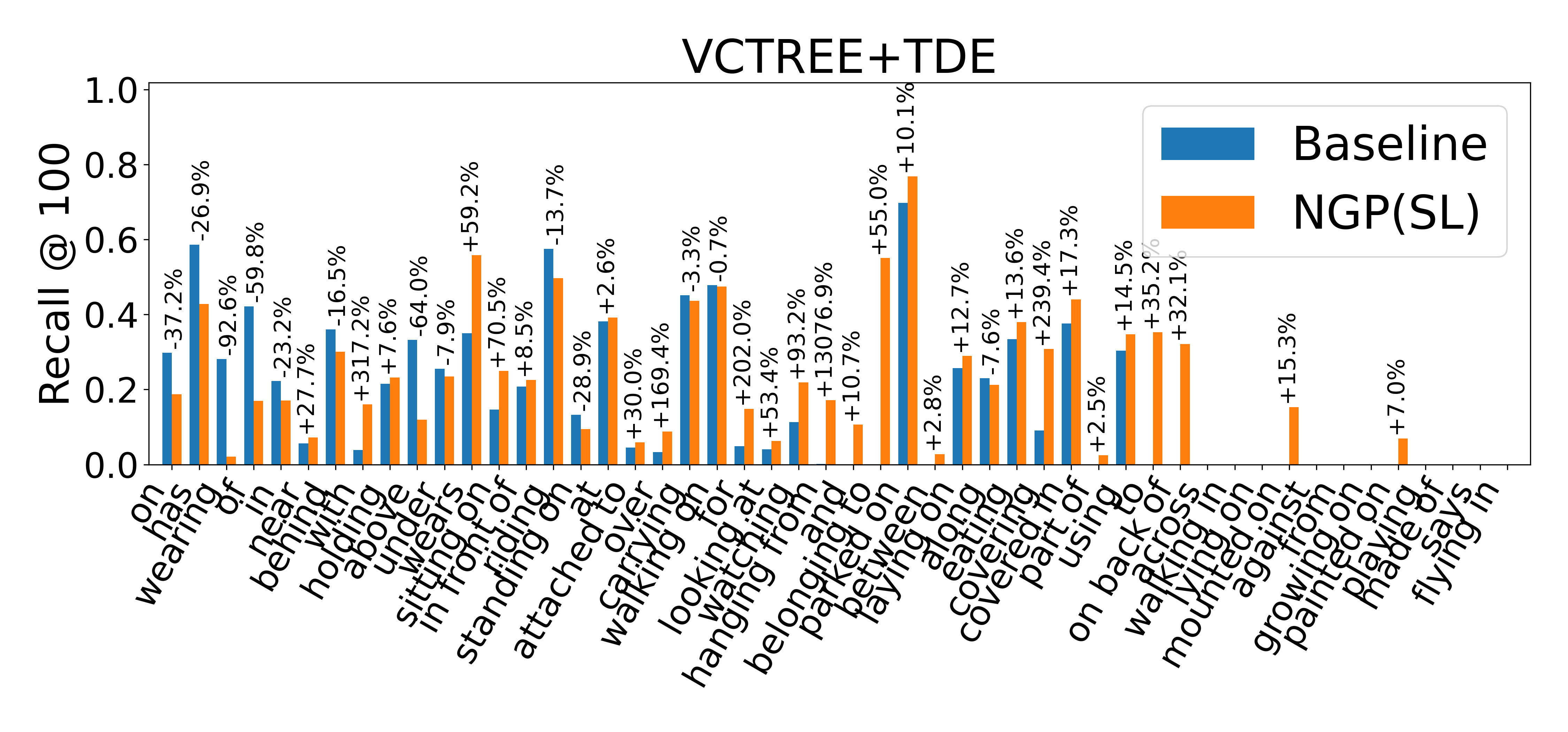} \\
\end{tabular}
\caption{Recall for scene graph classification on a per-predicate basis. Results on the VG dataset.}
\label{Figure: Empirical Results 4}
\end{figure*}

Figures~\ref{Figure: Empirical Results 3} and \ref{Figure: Empirical Results 4}
show results on {mR@100} on a per-predicate basis for predicate and scene graph classification, classification. The benchmark is VG.
Regularization is performed using NGP(SL) with $\theorya$ and $\rho=2$.
The relative recall improvements can be substantial. 
NGP can be seen as a form of weak supervision, where the model is provided with feedback that comes from ICs encoding commonsense knowledge in addition to the training signal that comes from the image labels. Intuitively, this signal can be seen as a way to help the model in better learning the less frequent predicates as for the more frequent ones there is already plenty of signal from the labelled data. A model trained only with the ground-truth would lack this signal.

It is worth noting that the baseline recall may drop for the frequent predicates in some cases.
This is due to inaccuracies in the ground truth \cite{TDE}
making the vast majority of the ground truth facts erroneously referencing only very few predicates, see Figure~\ref{Figure: Empirical Results 1}. For instance, in quite a few cases, the ground truth facts use the predicate ``on’’, even when ``laying on'' better describes the relationship between the subject and the object. Due to inaccuracies in the ground truth, the recall of a model that is more effective in detecting less frequent predicates, may be erroneously reported lower for the more frequent predicates.

\end{document}